\documentclass[10pt,twocolumn,letterpaper]{article}

\usepackage[pagenumbers]{cvpr} 

\usepackage[dvipsnames]{xcolor}

\definecolor{cvprblue}{rgb}{0.21,0.49,0.74}
\usepackage[pagebackref,breaklinks,colorlinks,citecolor=cvprblue]{hyperref}
\usepackage{mathtools}
\usepackage{amsmath}
\usepackage{multirow}
\usepackage{wrapfig}
\usepackage{amssymb}
\usepackage{pifont}
\usepackage{color,colortbl}
\usepackage{graphicx}
\usepackage{caption}
\definecolor{ggray}{rgb}{0.96, 0.96, 0.99}
\definecolor{lightgray}{rgb}{0.98, 0.98, 0.99}

\newcommand\norm[1]{\lVert#1\rVert}

\def\paperID{*****} 
\def\confName{CVPR}
\def\confYear{2024}

\title{LaMamba-Diff: Linear-Time High-Fidelity Diffusion Models Based on \\ Local Attention and Mamba}

\author{%
  Yunxiang Fu
  \ \ \ \
  Chaoqi Chen 
  \ \ \ \
  Yizhou Yu \\
  The University of Hong Kong
  	\\
	{\tt \small yunxiang@connect.hku.hk}, {\tt \small cqchen1994@gmail.com},  {\tt \small yizhouy@acm.org}
}

\begin{document}

\twocolumn[{%
\renewcommand\twocolumn[1][]{#1}%
\maketitle
\vspace{-8mm}
\begin{center}
    \centering
    \captionsetup{type=figure}
    \includegraphics[width=1\linewidth]{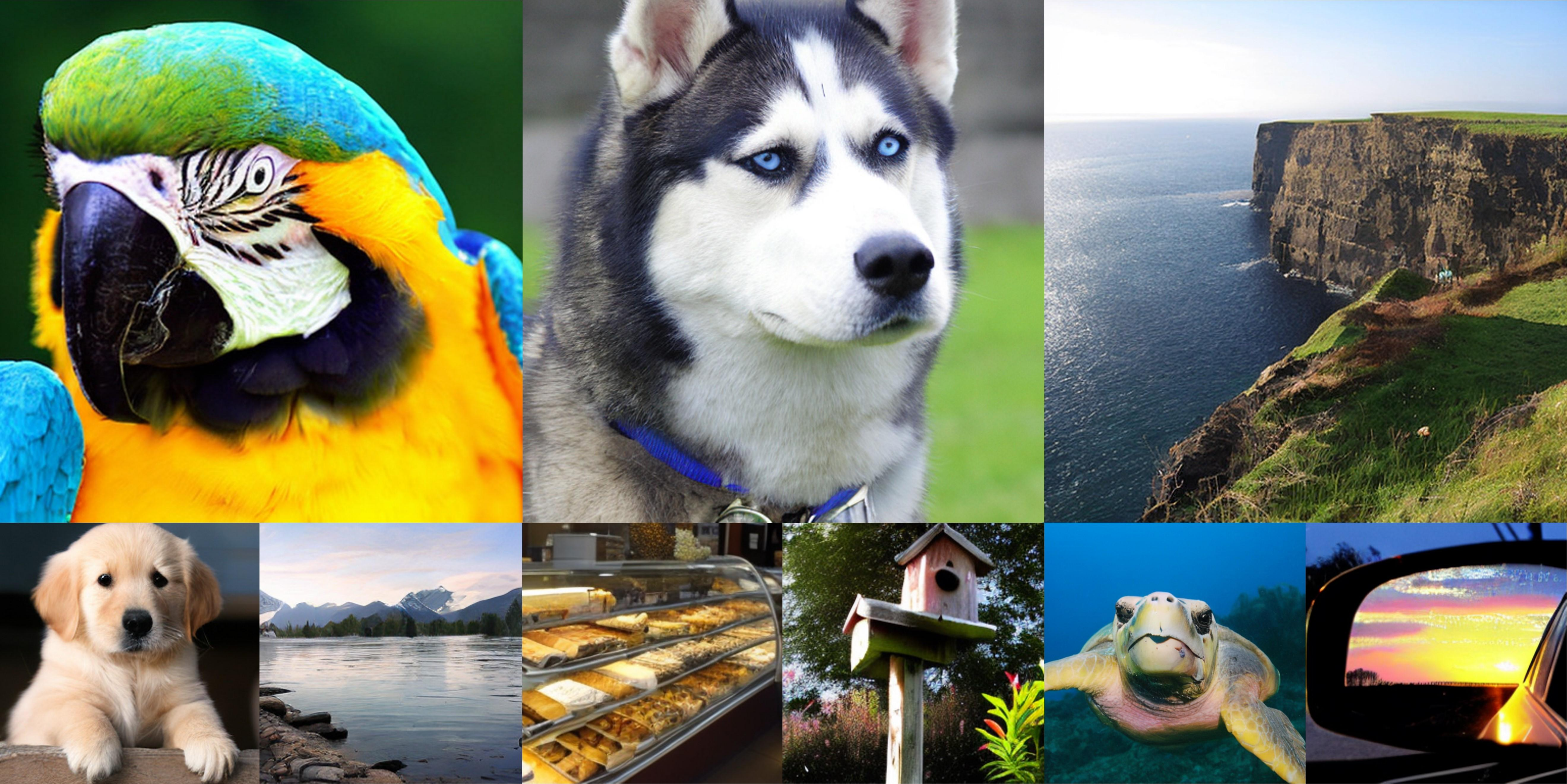}
    \vspace{-6mm}
    \caption{Sample images generated by our model trained on ImageNet at 512×512 and 256×256 resolutions.}
    \label{main_samples}
\end{center}%
}]

\begin{abstract}
\vspace{-5mm}
Recent Transformer-based diffusion models have shown remarkable performance, largely attributed to the ability of the self-attention mechanism to accurately capture both global and local contexts by computing all-pair interactions among input tokens.
However, their quadratic complexity poses significant computational challenges for long-sequence inputs. 
Conversely, a recent state space model called Mamba offers linear complexity by compressing a filtered global context into a hidden state. Despite its efficiency, compression inevitably leads to information loss of fine-grained local dependencies among tokens, which are crucial for effective visual generative modeling.
Motivated by these observations, we introduce Local Attentional Mamba (LaMamba) blocks that combine the strengths of self-attention and Mamba, capturing both global contexts and local details with linear complexity. 
Leveraging the efficient U-Net architecture, our model exhibits exceptional scalability and surpasses the performance of DiT across various model scales on ImageNet at 256x256 resolution, all while utilizing substantially fewer GFLOPs and a comparable number of parameters.
Compared to state-of-the-art diffusion models on ImageNet 256x256 and 512x512, our largest model presents notable advantages, such as a reduction of up to 62\% GFLOPs compared to DiT-XL/2, while achieving superior performance with comparable or fewer parameters. Our code is available at \href{https://github.com/yunxiangfu2001/LaMamba-Diff}{https://github.com/yunxiangfu2001/LaMamba-Diff}.
\end{abstract}

\vspace{-3mm}

\begin{figure*}[t]
\centering
    \begin{minipage}[t]{0.45\textwidth}
\centering
\vspace{0.2cm}
\makeatletter\def\@captype{figure}
\includegraphics[width=\textwidth]{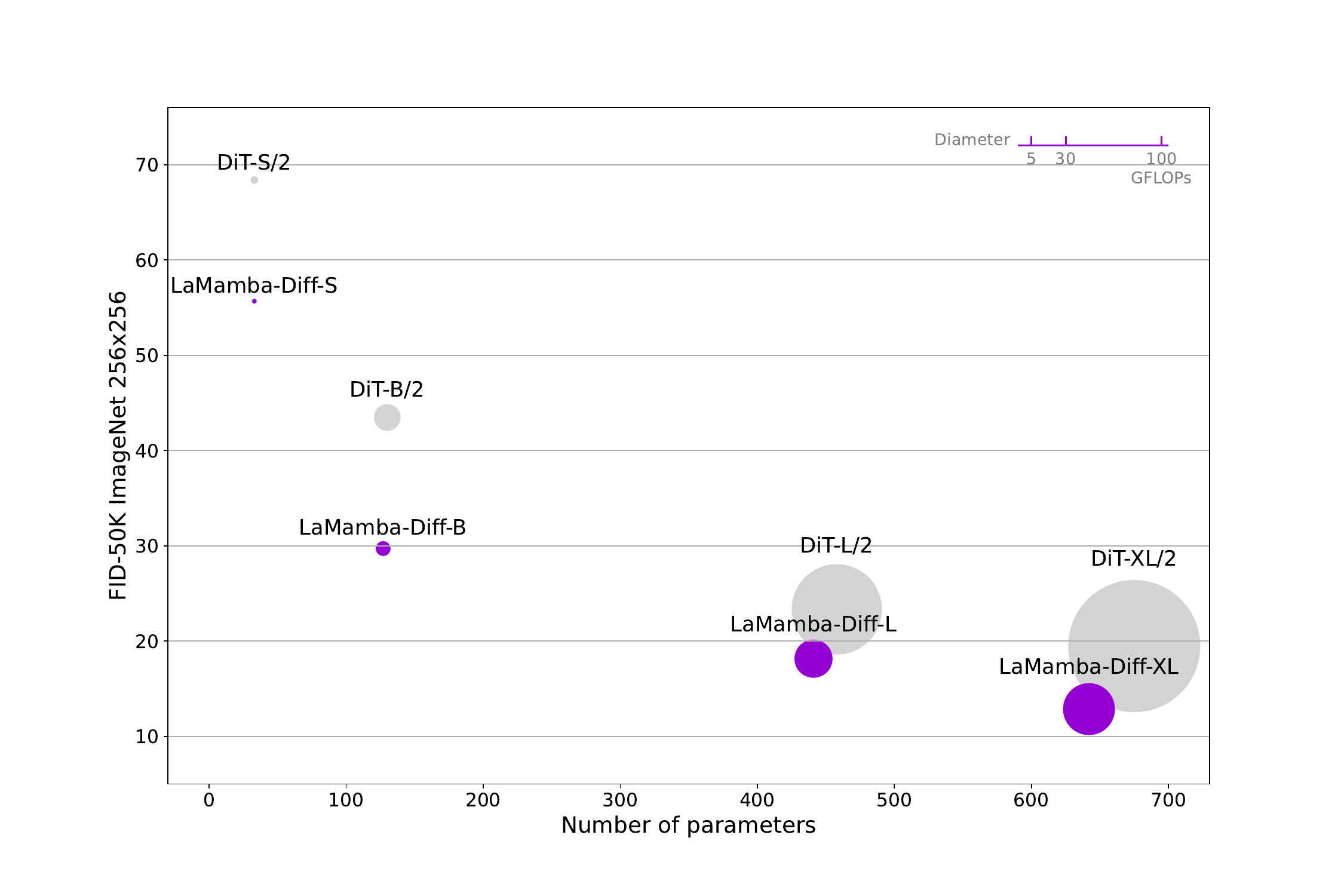}
\label{scaling}
\end{minipage}
\quad
\begin{minipage}[t]{0.45\textwidth}
\centering
\vspace{0.15cm}
\makeatletter\def\@captype{figure}

    \includegraphics[width=\textwidth]{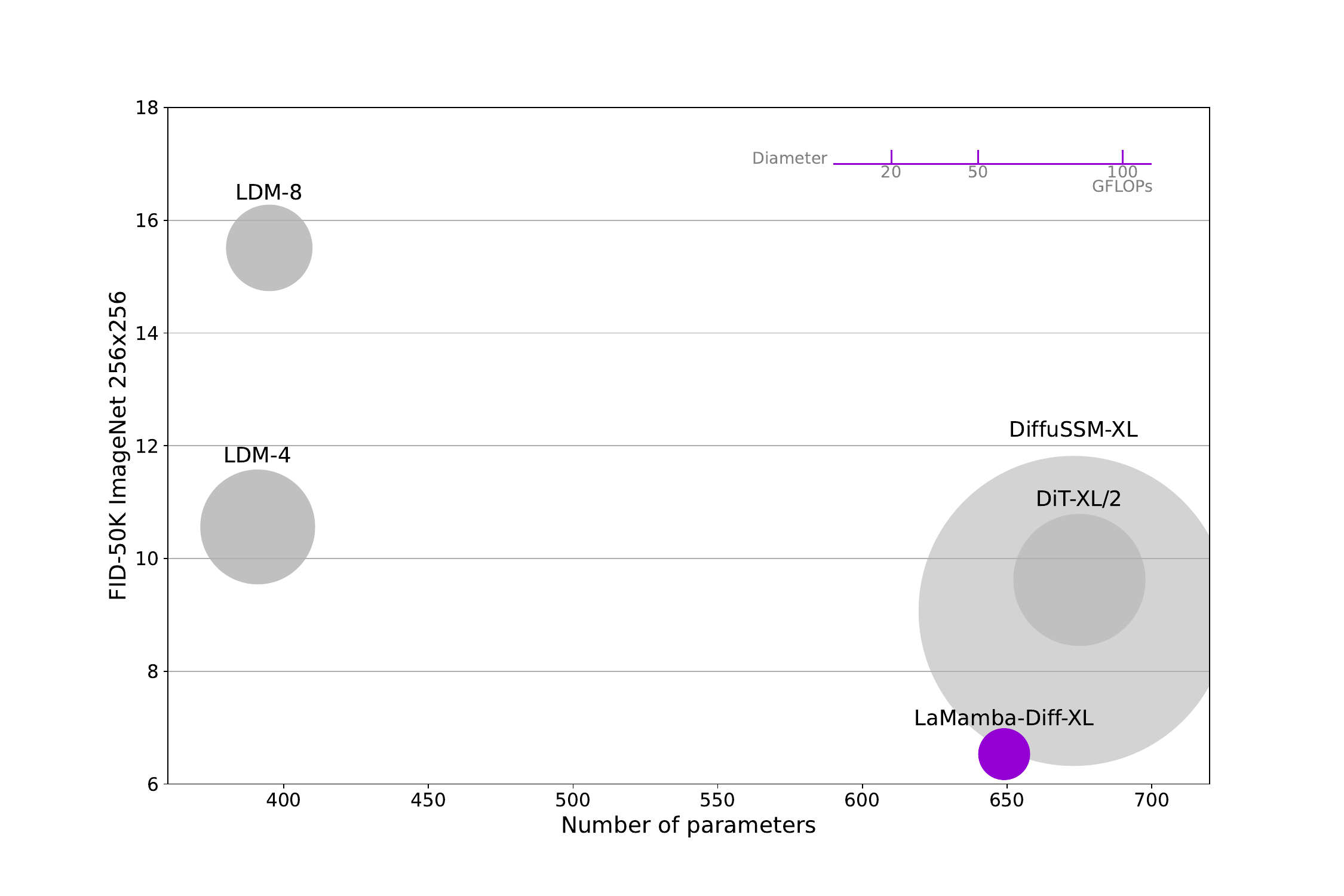}

\label{sota_diffusion_models}
\end{minipage}
\vspace{-4mm}
\caption{Unconditional image generation quality on ImageNet 256x256. The area of bubbles denote GFLOPs. \textit{Left}: FID-50K of LaMamba-Diff models trained for 400k iterations. Performance improves with the number of parameters and GFLOPs. \textit{Right}: Our largest model outperforms state-of-the-art diffusion models with substantially fewer GFLOPs.}
\label{intro_FID_comparisons}
\end{figure*}

\section{Introduction}
Diffusion models have made significant strides in the field of generative modeling~\cite{ho2020denoising,song2019generative}, especially for images~\cite{dhariwal2021diffusion,rombach2022high,saharia2022photorealistic,nichol2021glide,ramesh2021zero,ramesh2022hierarchical}, often surpassing the previously popular generative adversarial networks (GANs)~\cite{goodfellow2014generative}.
The advancements of diffusion models have been driven by many orthogonal factors, including sampling approaches~\cite{ho2020denoising,karras2022elucidating,ho2022classifier,song2020denoising}, latent space modeling~\cite{rombach2022high}, and backbone architecture designs~\cite{peebles2023scalable,yan2023diffussm}. 
Despite recent attempts to devise innovative backbones, state-of-the-art backbones for diffusion models~\cite{rombach2022high,peebles2023scalable,jabri2022scalable} largely rely on self-attention~\cite{vaswani2017attention} for high-fidelity generation. The effectiveness of self-attention stems from its capacity to accurately capture global contexts and fine-grained local dependencies from inputs by explicitly computing all-pair interactions among input tokens. However, their quadratic computational complexity with respect to the input sequence length poses computational challenges when dealing with long sequences, such as high-resolution images or sequences formed with a small patch size.

To address the quadratic complexity of self-attention, a recent state space model (SSM) named Mamba~\cite{gu2023mamba} has been proposed. Using dynamic weights, Mamba captures global contexts with linear time by compressing contextual information of input tokens into a hidden state.
Demonstrating a strong potential in long-sequence modeling, Mamba has been applied across diverse domains, including medical data~\cite{liu2024swin,ruan2024vm}, point clouds~\cite{liu2024point}, and vision tasks~\cite{zhu2024vim,liu2024vmamba}. However, unlike self-attention, the compression and selective process of Mamba does not explicitly compute pairwise interactions. 
Consequently, part of fine-grained local information is lost, leading to sub-optimal performance. The significance of local details for generative modeling is demonstrated by the detrimental decline in the performance of DiT~\cite{peebles2023scalable} as the patch size increases, attributed to the loss of fine-grained information within each patch~\cite{han2021transformer}.

In light of these observations, we introduce a novel Local Attentional Mamba (LaMamba) block that combines the strengths of Transformers and Mamba to accurately model global contexts and local details with linear complexity. LaMamba captures global contexts efficiently using Mamba while accurately preserving fine-grained local dependencies using local self-attention. Our local self-attention has linear complexity by computing pairwise interactions within a context window with a fixed size. LaMamba brings the best of two worlds for visual generative modeling, offering favorable properties like scalability, robustness, and efficiency.

Based on LaMamba blocks, we design LaMamba-Diff, a novel backbone for diffusion models that adopts a U-Net architecture~\cite{ronneberger2015u}. LaMamba-Diff naturally constructs multi-scale hierarchical features through down and up-sampling and exhibits efficiency by compressing spatial dimensions during the downsampling phase. 
The highly efficient design of LaMamba and LaMamba-Diff allows us to utilize \(1\times 1\) patches, which enables more accurate modeling of fine-grained spatial dependencies by preventing the loss of local details within each patch~\cite{han2021transformer}. In particular, with a comparable number of parameters, LaMamba-Diff with a patch size of 1 uses significantly fewer GLFOPs compared to DiT~\cite{peebles2023scalable} with a patch size of 2, despite that DiT has a 4x shorter input sequence length.

We evaluate the performance of LaMamba-Diff on the widely-used ImageNet dataset for image generation. As illustrated in Fig.~\ref{intro_FID_comparisons} (right), LaMamba-Diff achieves state-of-the-art Fréchet Inception Distance (FID) with comparable or fewer parameters and significantly less GFLOPs on ImageNet at \(256\times 256\) resolution. 
Moreover, as depicted in Fig.~\ref{intro_FID_comparisons} (left), LaMamba-Diff exhibits excellent scalability with FID scores consistently decreasing with the number of parameters and GFLOPs.
In the class-conditional ImageNet 256x256 image generation benchmark, LaMamba-Diff-XL achieves a state-of-the-art FID of 2.04 using 57.6\% fewer GFLOPs compared to DiT-XL/2. For class-conditioned ImageNet 512x512 image generation, LaMamba-Diff-XL achieves an FID of 3.01 using 61.6\% fewer GFLOPs compared to DiT-XL/2. 
These results demonstrate the effectiveness and efficiency of LaMamba-Diff for high-resolution image generation tasks.

To summarise, our contributions in this paper are three-fold:
\begin{itemize}
    \item We design a novel LaMamba block that combines the strengths of Transformers and Mamba, accurately capturing global contexts and local details with linear complexity.
    \item We propose LaMamba-Diff, a highly efficient backbone for diffusion models. LaMamba-Diff allows a patch size of 1 without substantial computational overhead, which was previously unavailable for diffusion backbones.
    \item Experimental results demonstrate that LaMamba-Diff achieves state-of-the-art FID using significantly fewer GFLOPs. Additionally, LaMamba-Diff demonstrates excellent scalability.

\end{itemize}

\section{Related works}
\textbf{Diffusion model backbones.}
Diffusion models (DM) belong to a class of probabilistic generative models that iteratively corrupt data by introducing noise through a forward process, and subsequently learn to reverse this process for sample generation~\cite{ho2020denoising,song2020score,sohl2015deep}. Recently, DM has become the de facto choice for image generation, due to its capability to generate photo-realistic images and stable training property~\cite{saharia2022photorealistic,rombach2022high,ramesh2021zero,rombach2022high,chang2023muse,ramesh2021zero,yang2022diffusion}, which was unavailable for the previous state-of-the-art GANs~\cite{goodfellow2014generative}. The backbones for state-of-the-art DM are UNet-based~\cite{ho2020denoising,rombach2022high} or Transformer-based~\cite{peebles2023scalable} and rely heavily on attention~\cite{vaswani2017attention}. However, the quadratic complexity of attention hinders the application of diffusion models to long sequential data. Very recently, to achieve linear complexity, state space models have been leveraged to construct backbones for DM~\cite{yan2023diffussm,fei2024DiS,hu2024zigma} by replacing the self-attention module in existing backbones like DiT~\cite{peebles2023scalable} with Mamba~\cite{gu2023mamba}. However, they fail to outperform DiT using a comparable number of parameters and GFLOPs. 
In contrast, we propose a hybrid Mamba backbone using local attention, achieving linear complexity without sacrificing image fidelity and outperforming DiT using fewer GFLOPs and parameters.

\textbf{State space models.}
A recent state space model named Mamba~\cite{gu2023mamba} has gained popularity in sequence modeling renowned for its linear complexity, dynamic weights, and global receptive field.
Mamba has been widely explored in various domains, including vision backbones~\cite{zhu2024vim,liu2024vmamba}, medical imaging~\cite{liu2024swin,ruan2024vm}, 3D point clouds~\cite{liu2024point}, tabular data~\cite{ahamed2024mambatab}, and image/video generation~\cite{yan2023diffussm,fei2024DiS,hu2024zigma,oshima2024ssm}.
In this paper, we investigate the use of Mamba for image generation, distinguishing ourselves from previous works by utilizing local attention to address the issue of fine-grained local detail loss when replacing attention with Mamba. 
Concurrently, the Matten model~\cite{gao2024matten} explores different combinations of Mamba and attention for video generation using an isotropic architecture, but exhibits quadratic complexity with sequence lengths. 
In contrast, our work has linear complexity and employs an efficient U-Net architecture that permits a patch size of 1, thereby avoiding the loss of local information within patches~\cite{han2021transformer}.

\section{Preliminary}

\textbf{Latent diffusion models (LDMs)}~\cite{rombach2022high} typically operate on the latent space of a pre-trained variational autoencoder (VAE)~\cite{kingma2013auto,esser2021taming}. The pre-trained VAE encoder \(\mathcal{E}\) is used to transform data samples \(x_0 \sim p_{data}(x)\) into latent representation \(z_0=\mathcal{E}(x_0)\) with a downsampling factor of 8. LDMs learn to sample from this latent distribution by progressively perturbing \(z\) in a forward process, and then learn to reverse this process. Specifically, forward diffusion gradually convert \(z_0\) to a prior noise distribution with intermediate noisy latents \(z_1, ..., z_T\) and \(z_T \sim \mathcal{N}(0,\pmb{I})\) by applying Gaussian transitions: \(
q(z_t|z_{t-1}) = N(z_t;\sqrt{1-\beta_t}z_{t-1}, \beta_t\pmb{I}),
\)
where \(t \in \{1,...,T\}\) denote the timestep and \(\beta_t \in (0,1)\) is a predefined noise variance hyperparameter. 

LDMs are trained to learn the reverse diffusion process 
\(
p_{\theta}(z_{t-1}|z_t) = \mathcal{N}(z_{t-1};\mu_{\theta}(z_t),\Sigma_{\theta}(z_t) ),
\) where mean \(\mu_{\theta}(z_t)\) and variance \(\Sigma_{\theta}(z_t)\) are parameterized by neural networks. In practice, the variance \(\Sigma_{\theta}\) can be predefined or learned, and the mean \(\mu_{\theta}\) is parameterized using a noise predictor \(\epsilon_{\theta}\)~\cite{ho2020denoising,nichol2021improved}. The training loss can be rewritten as a simplified form of the variational lower bound~\cite{kingma2013auto} \(\mathcal{L}_{simple} = \sum^{T-1}_{t=1} \mathbb{E}_{z_0, \epsilon}[\norm{\epsilon_t - \epsilon_{\theta}(z_t,t)}^2]\)~\cite{nichol2021improved}. 
Additionally, we follow the common practice to train \(\Sigma_{\theta}\) using \(\mathcal{L}_{vlb}\)~\cite{peebles2023scalable,nichol2021improved}.
After training, novel latents \(z_0\) can be sampled by iteratively denoising a random noise \(z_T\) using the noise predicted by \(\epsilon_{\theta}\) with \(z_{t-1} \sim p_{\theta}(z_{t-1}|z_t)\) via the reparameterization trick.
Subsequently, images are generated by passing \(z_0\) to the pre-trained VAE decoder. 
We follow LDMs by training diffusion models in the latent space of a pre-trained VAE.

\textbf{State space models (SSM)}~\cite{gu2021efficiently,gu2021combining,gu2023mamba,smith2022simplified} are sequence-to-sequence models that map input signals \(x(t) \in \mathbb{R}\) to output signals \(y(t) \in \mathbb{R} \) through a latent \(h(t) \in \mathbb{R}^{N\times 1}\). The continuous time process for a linear time-invariant (LTI) SSM can be formulated as 
\begin{align}
    \begin{split}
        h'(t) &= Ah(t) + Bx(t),\\
        y(t) &= Ch(t) + \Bar{D}x(t), 
    \end{split}
\label{continuous_ssm}
\end{align}
where \(A \in \mathbb{R}^{N\times N}\), \(B \in \mathbb{R}^{N\times 1}, C\in \mathbb{R}^{1\times N}\) denote the diagonal state matrix, input matrix, and output matrix for hidden size \(N\), respectively, and \(\Bar{D} \in \mathbb{R}\) is a shortcut that provides a direct path from input to \(y\).

In order to utilize SSM with real-world data, which is typically discrete, the continuous process represented in Equation (\ref{continuous_ssm}) is discretized using the zero-order hold (ZOH) rule, which is formulated as~\cite{gupta2022diagonal}
\begin{align}
    \begin{split}
        h_t &= \Bar{A}h_{t-1} + \Bar{B}x_t,\\
        y_t &= Ch_{t} + \Bar{D}x_t, 
    \end{split}
\end{align}
where \(\Bar{A} = e^{(\Delta A)} \) and \(\Bar{B} = (\Delta A)^{-1}(e^{(\Delta A)}-I)\cdot (\Delta B) \) with \(\Delta\) as the timescale parameter for ZOH. 

Mamba~\cite{gu2023mamba} relaxes the LTI constraint and introduces a selective scan mechanism (S6) that allows hidden states \(h\) to be contextually aware of the input using dynamic weights. Specifically, S6 introduces selectivity to \(\Delta, B\), and \(C\), enabling them to capture the context of input tokens, while the context of \(\Bar{A}\) is obtained from \(\Delta\). Given an input \(x_t \in \mathbb{R}^D\) with \(D\) channels, S6 independently selects the context of each channel, and compresses the selected context for each channel into a hidden state \(h \in \mathbb{R}^N\), with \(N\) representing the SSM state dimension. 
Our proposed model builds on the strengths of Mamba, leveraging its ability to capture global contextual information in linear time, while also addressing the loss of fine-grained details through the incorporation of local attention.


\section{Methodology}
\begin{figure*}[t]
\centering
\includegraphics[width=0.98\textwidth]{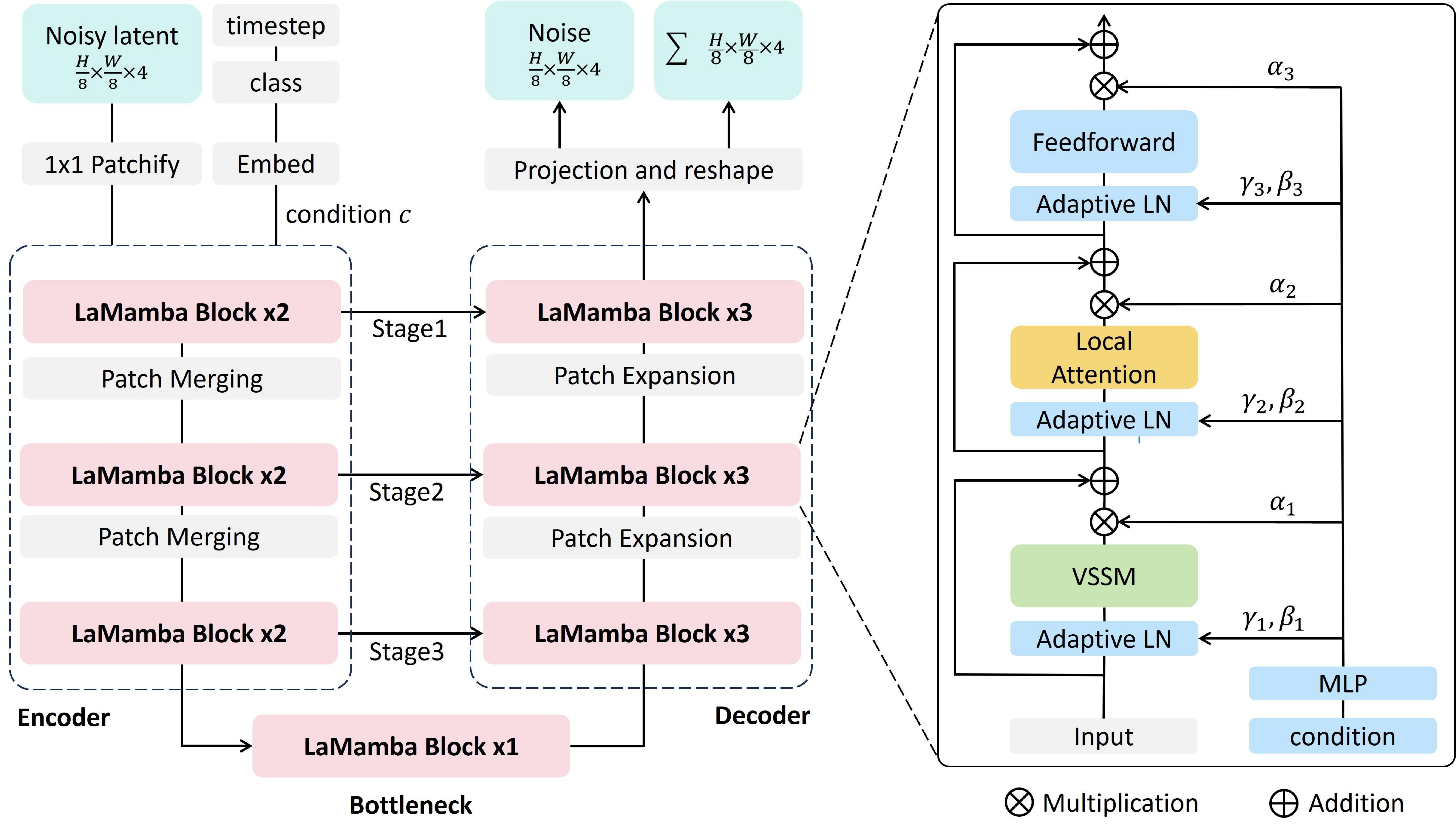}
\caption{Network architecture of LaMamba-Diff. \textit{Left:} Architecture of LaMamba-Diff-S. \textit{Right:} Local attentional Mamba block.}
\label{method_fig}
\end{figure*}

We present a novel linear-time hybrid U-Net architecture named Local Attentional Mamba for Diffusion (LaMamba-Diff). LaMamba-Diff, built upon Local Attentional Mamba (LaMamba) blocks, is designed to combine the strengths of Mamba and self-attention, having dynamic weights that efficiently capture global contexts while precisely modeling local dependencies.  
Fig.\ref{method_fig} provides an illustration of LaMamba-Diff. 

\subsection{LaMamba blocks}
\label{method_LaMamba}
\textbf{Overview.}
Transformers have demonstrated remarkable performance as backbones in the latent diffusion framework~\cite{rombach2022high,peebles2023scalable}. The effectiveness of Transformers can be attributed to self-attention, which is capable of capturing detailed contextual information from all input token pairs. However, the quadratic complexity of self-attention with respect to the length of the input sequence poses scalability challenges for long input sequences arising from high-resolution scenarios or small patch sizes.
In contrast, Mamba~\cite{gu2023mamba} recurrently compresses contextual information of an input sequence into a single hidden state to achieve linear time complexity. However, this approach inevitably suffers from a loss of detailed information in input tokens due to compression. However, such detailed information is crucial to visual generative modeling.
This limitation is evident in performance comparisons between SSM-based diffusion models and DiT, where the former fails to outperform the latter when using a comparable number of parameters and GFLOPs~\cite{yan2023diffussm,fei2024DiS,peebles2023scalable}.
To bring the best of both worlds, we introduce Local Attentional Mamba (LaMamba) blocks that capture both global contexts and local details in linear time.

Our proposed LaMamba block is illustrated in Fig.~\ref{method_fig} (right) and consists of three components: a visual state space model (VSSM), a local attention module, and a feedforward network (FFN). 
Every component is equipped with a residual connection. Additionally, every component integrates condition information, such as timesteps and class labels, using adaptive layer normalization (AdaLN)~\cite{dhariwal2021diffusion} and a scaling operation before the residual connection.
The VSSM is a variation of the original Mamba block tailored for 2D visual data~\cite{liu2024vmamba} and is used to efficiently capture global contexts. We adopt spatially continuous scans in four complementary directions~\cite{yang2024plainmamba}.
Local attention explicitly computes local pairwise attentions among input tokens without compression, resulting in fine-grained local representations. 
The standard FFN with a hidden dimension expansion ratio of 4 is added at the end of the block.
LaMamba integrates conditions using AdaLN, where the scale and shift parameters ($\gamma$ and $\beta$) of Layer Normalization~\cite{ba2016layer} are regressed from the condition embedding using a multi-layer perceptron (MLP).
Additionally, a dimension-wise scaling operation is adopted for zero-initialization~\cite{goyal2017accurate,peebles2023scalable}, where the MLP is initialized to make the scaling parameter ($\alpha$) a zero-vector at the start of training.
This makes the LaMamba block an identity function initially as the output of each component is equal to its residual connection.
As the visual state space model (VSSM), local attention module, and feed-forward network (FFN) all exhibit linear time complexity, LaMamba also maintains linear time complexity. Furthermore, the computational cost associated with adaptive layer normalization (AdaLN) and zero-initialization is negligible in terms of GFLOPs.


\textbf{Main components.}
Originally designed for sequence modeling in natural language processing, Mamba casually processes 1D input sequences. Directly applying Mamba to non-casual 2D visual signals may yield sub-optimal results as it does not consider 2D spatial information, which is crucial for vision tasks~\cite{zhu2024vim,liu2024vmamba}. 
Hence, we adopt a visual state space module (VSSM) that explicitly models 2D spatial information. We explore four variants of VSSM: (1) Bi-directional Mamaba (ViM)~\cite{zhu2024vim} adds positional embeddings and performs both forward and backward scanning to incorporate spatial information.
(2) LocalVMamba~\cite{huang2024localmamba} partitions input tokens into windows, each of which is scanned individually to ensure local spatial information is encoded closely together. Afterwards, a global scan is carried out across windows.
(3) EfficientVMamba~\cite{pei2024efficientvmamba} implements multiple scanning paths efficiently by scanning tokens with a fixed step size larger than one and aggregating the results along different scanning paths. This approach allows tokens to integrate information from multiple scanning directions without introducing computational overhead.
(4) 2D Selective-Scan (SS2D) blocks~\cite{liu2024vmamba} adopt four distinct and complementary scanning paths, enabling each token to integrate information from all other tokens in four different directions. Moreover, SS2D removes the multiplicative
branch in original Mamba blocks~\cite{gu2023mamba}. We modify the scanning trajectories in SS2D to spatial continuous scans~\cite{yang2024plainmamba}. Unless otherwise stated, all models in this work use our modified SS2D block as the VSSM owing to its superior performance, as shown in the ablation study presented in Table~\ref{ablation_mamba}.


The compression and selection mechanism in VSSM serves to filter out irrelevant information for improved efficiency~\cite{gu2023mamba} but may lose information about fine-grained local details and dependencies.
To address this issue, we incorporate a local attention module for fine-grained local representations. 
In practice, we utilize window-based attention~\cite{liu2021swin}, where the input is partitioned into non-overlapping square windows and self-attention is computed within each window to explicitly capture local dependencies without compression.
Note that complexity is linear with respect to sequence length because the size of every window is fixed and hence the number of tokens within every window is a constant.

To integrate conditions into diffusion models, we follow the widespread usage of AdaLN~\cite{peebles2023scalable,dhariwal2021diffusion}. Concretely, we replace layer normalization~\cite{ba2016layer} with AdaLN by regressing the scale and shift parameters (\(\gamma\) and \(\beta\)) using the sum of the embedding vectors of timestep \(t\) and class label \(l\). This process can be formulated as \(\text{AdaLN}(h,c) = \gamma \text{LN}(h) + \beta\), where \(\gamma\) and \(\beta\) are computed by passing the conditioning embedding \(c\) to an MLP. 

\textbf{Summary.} The novelty of LaMamba lies in the composition of Mamba and local attention modules to complement each other within each block.
Despite that LaMamba is based on VSSM~\cite{liu2024vmamba} and windowed local attention~\cite{liu2021swin}, to the best of our knowledge, this is the first attempt to incorporate both Mamba and local attention within the building block of a diffusion model backbone, resulting in the capability to capture both global contexts and fine-grained local details with linear time complexity.
LaMamba naturally inherits the strengths of both Transformer and Mamba, including scalability, robustness, and efficiency. This is demonstrated by the state-of-the-art performance and scalability analysis on ImageNet, as shown in Table~\ref{main_imagenet} and Figure~\ref{intro_FID_comparisons}, respectively.
Moreover, ablation studies in Table~\ref{ablation_attention} empirically confirm the effectiveness of local attention in learning fine-grained local representations.


\newcommand\Tstrut{\rule{0pt}{2.6ex}}
\newcommand\Bstrut{\rule[-0.9ex]{0pt}{0pt}}
\begin{table*}[t]
\small
\centering
\setlength\tabcolsep{2pt}
\scalebox{0.9}{
\begin{tabular}{l|cc|c@{\hskip 6pt}|@{\hskip 6pt}c@{\hskip 6pt}|@{\hskip 6pt}c@{\hskip 6pt}|@{\hskip 6pt}c@{\hskip 6pt}} 
\toprule
 & Stage &  {Output size} & Small & Base & Large & X-Large\\
\midrule
\midrule
\multirow{6}{*}{Encoder} & \multirow{2}{*}{1} & \multirow{2}{*}{\(16\times 16\)} & \hspace{2pt} [LaMamba] \(\times 2\) (96) & [LaMamba] \(\times 2\) (192) & [LaMamba] \(\times 2\) (256) & [LaMamba] \(\times 2\) (320) \\
 &  &  & \hspace{2pt} Patch Merge (192) & Patch Merge (384) &  Patch Merge (512) &  Patch Merge (640) \\[1.5mm]
 \cline{2-7}
 
 & \multirow{2}{*}{2} & \multirow{2}{*}{\(8\times 8\)} & \hspace{2pt} [LaMamba] \(\times 2\) (192) & [LaMamba] \(\times 2\) (384) & [LaMamba] \(\times 2\) (512) & [LaMamba] \(\times 2\) (640) \Tstrut\\
 &  &  & \hspace{2pt} Patch Merge (384) & Patch Merge (768) &  Patch Merge (1024) &  Patch Merge (1280)\\[1.5mm]
 \cline{2-7}
 
 & 3 & {\(8\times 8\)} & \hspace{2pt} [LaMamba] \(\times 2\) (384) & [LaMamba] \(\times 2\) (768) & [LaMamba] \(\times 2\) (1024) & [LaMamba] \(\times 2\) (1280) \Tstrut\\[1mm]
 \cline{2-7}
 & 4 & {\(8\times 8\)} & \hspace{2pt} Identity (384) & Identity (768) & [LaMamba] \(\times 2\) (1024) & [LaMamba] \(\times 2\) (1280) \Tstrut\\

\midrule
Bottleneck & - & \(8\times 8\) & \hspace{2pt} [LaMamba] \(\times 1\) (384) & [LaMamba] \(\times 1\) (768) & [LaMamba] \(\times 2\) (1024) & [LaMamba] \(\times 2\) (1280) \\
\midrule

\multirow{6}{*}{Decoder}  & 4 & {\(8\times 8\)} & \hspace{2pt} Identity (384) & Identity (768) & [LaMamba] \(\times 3\) (1024) & [LaMamba] \(\times 3 \) (1280) \\[1mm]
\cline{2-7}

 & 3 & {\(8\times 8\)} & \hspace{2pt} [LaMamba] \(\times 3\) (384) & [LaMamba] \(\times 3\) (768) & [LaMamba] \(\times 3\) (1024) & [LaMamba] \(\times 3\) (1280) \Tstrut\\[1mm]
 \cline{2-7}
 
 & \multirow{2}{*}{2} & \multirow{2}{*}{\(16\times 16\)} & \hspace{2pt} Patch Expand (192) & Patch Merge (384) &  Patch Merge (512) &  Patch Merge (640) \Tstrut\\
 & &  & \hspace{2pt} [LaMamba] \(\times 3\) (192) & [LaMamba] \(\times 3\) (384) & [LaMamba] \(\times 3\) (512) & [LaMamba] \(\times 3\) (640) \\[1.5mm]
 \cline{2-7}

  & \multirow{2}{*}{1} & \multirow{2}{*}{\(32\times 32\)} & \hspace{2pt} Patch Expand (96) & Patch Merge (192) &  Patch Merge (256) &  Patch Merge (320) \Tstrut\\
 & &  & \hspace{2pt} [LaMamba] \(\times 3\) (96) & [LaMamba] \(\times 3\) (192) & [LaMamba] \(\times 3\) (256) & [LaMamba] \(\times 3\) (320) \\
 
\midrule
Output layer & - & \(32\times 32\) & \multicolumn{4}{c}{adaLN -> Linear (8)}\\
\midrule

Param & - & - & 32M & 127M & 449M & 656M \\
GFLOPs & - & - & 3.19 & 12.32 & 33.39 & 49.90 \\
\bottomrule
\end{tabular}}
\caption{Summary of LaMamba-Diff network architectures. The noisy latent produced from all our networks is shaped \(32\times 32\), which corresponds to 256x256 input images. The hidden dimensions of each block are shown in brackets.\vspace{-5mm}}
\label{architecture_overview}
\end{table*}

\subsection{Network architecture}
\label{method_architecture}
The inherent hierarchical structure observed in natural images~\cite{saremi2013hierarchical} motivates the adoption of hierarchical models for image generation~\cite{ho2020denoising,rombach2022high}.
This approach has proven successful in the U-Net architecture, which naturally constructs multi-scale hierarchical representations, commonly used for diffusion models~\cite{ronneberger2015u,ho2020denoising,rombach2022high}, but not for diffusion Transformers~\cite{peebles2023scalable,bao2023one}. To leverage hierarchical features and improve efficiency, we adopt the U-Net architecture. Figure~\ref{method_fig}~(left) illustrates the architecture of LaMamba-Diff-S. In particular, we follow LDMs with LaMamba-Diff operating in the latent space of the pre-trained VAE provided by Stable Diffusion~\footnote{https://huggingface.co/stabilityai/sd-vae-ft-mse}~\cite{rombach2022high}, which has a down-sampling factor of 8. 

\textbf{Input tokens.} The input of LaMamba-Diff consists of timestep \(t\), class label \(l\), and noisy latent \(z_t \in \mathbb{R}^{\frac{H}{8} \times \frac{W}{8} \times 4}\), which has the same dimensions as the latent of the pre-trained VAE.
By viewing \(z_t\) as a \(\frac{H}{8} \times \frac{W}{8}\) grid of features, we treat every position on the grid as a patch (i.e., \(1\times 1\) patch size) and linearly embed every patch without the need to flatten it into a 1-D sequence. This encodes \(z_t\) into a latent representation of size \(\frac{H}{8} \times \frac{W}{8} \times D\), where \(D\) is the hidden dimension.
Meanwhile, \(t\) and \(l\) are encoded into continuous representations using an MLP and a learnable embedding dictionary, respectively; the sum of these two encoding results becomes the conditioning embedding \(c\).

\textbf{U-Net architecture.} As illustrated in Fig.~\ref{method_fig} (left), LaMamba-Diff consists of an encoder, a bottleneck stage, and a decoder, with skip connections between corresponding stages of the encoder and decoder. 
Skip connections are implemented with channel-wise addition instead of channel concatenation.
Under similar number of parameters and computational cost (GFLOPs), skip connections with channel concatenation result in a smaller number of hidden dimensions, which can lead to sub-optimal performance as shown in the ablation study presented in Table~\ref{ablation_architecture}.
Hierarchical representations are built by downsampling feature maps from stages 1 and 2 of the encoder with patch merging~\cite{liu2021swin}, and upsampling feature maps fed into stages 1 and 2 of the decoder with patch expansion~\cite{cao2022swin}. Thus, the spatial resolution at the bottleneck stage is 4 times smaller than the input resolution of the encoder.


Notably, we allow the propagation of fine-grained local information across window boundaries in local attention by applying an alternating shifting scheme~\cite{liu2021swin} to every two consecutive blocks. Specifically, for every two blocks, the first block adopts the standard window partitioning scheme for local attention, while the second block adopts a windowing configuration that is shifted from the first block by \((\lfloor \frac{M}{2}\rfloor, \lfloor \frac{M}{2}\rfloor)\), where \(M\) represents the window size, which is set to 8 by default.

We design four distinct LaMamba-Diff models with different scales (Table~\ref{architecture_overview}), each having a comparable number of parameters to its corresponding DiT variant~\cite{peebles2023scalable}, allowing fair comparisons of the model architecture. The architecture overview has been presented in Section~\ref{method_architecture}. 
The condition embedding dimensions for Small, Base, Large, and X-Large are 192, 384, 1024, and 1280 respectively. 
All LaMamba-Diff variants have two downsampling and upsampling blocks only in stages 1 and 2. Our large and X-large variants have four stages without extra downsampling and upsampling blocks.

\textbf{Noise and covariance prediction.}
We apply AdaLN before linearly projecting and reshaping the output of the decoder to predict noise \(\epsilon\) and covariance \(\Sigma\),
which have the same dimensions as the input noisy latent \(z_t\) (i.e., \(\epsilon, \Sigma \in \mathbb{R}^{\frac{H}{8} \times \frac{W}{8} \times 4}\)). 

\textbf{Local detail preservation.} Fine-grained local information within each patch is crucial for excavating features from visual data~\cite{han2021transformer}. 
However, existing diffusion models based on Transformers (such as DiT~\cite{peebles2023scalable}) and SSMs (such as DiffuSSM~\cite{yan2023diffussm} and DiS~\cite{fei2024DiS}) do not effectively handle this aspect. 
When larger patch sizes are used, this limitation leads to a loss of local details and a corresponding increase in FID~\cite{peebles2023scalable}.
On the other hand, using smaller patch sizes results in an increased computational cost as the input sequence length is inversely proportional to the squared patch size.
We prevent the loss of valuable fine-grained information by utilizing \(1\times 1\) patches, which is computationally feasible thanks to the linear complexity and efficient design of LaMamba-Diff.

\subsection{Computational Complexity Analysis}
\label{method_computation_analysis}
In this section, we provide theoretical and empirical analysis of computational complexity for LaMamba-Diff. 
Since the selective scan 2D (SS2D) mechanism contributes to the majority of FLOPs in VSSM by computing 4 SSM processes with a hidden dimension expansion ratio of 2~\cite{liu2024vmamba}, we approximate the complexity of VSSM with SS2D.

Given a noisy diffusion latent \(z \in \mathbb{R}^{\frac{H}{8}\times \frac{W}{8} \times D}\), where \(D\) is the hidden dimension, the computational complexity of SS2D, windowed multi-head self-attention (W-MSA), and feedforward net (FFN)
are given below:
\begin{align}
        \Omega(\text{SS2D}) &= 4\Omega(\text{SSM}) \nonumber \\
        &=\; 4(3L(2D)N + L(2D)N), \\
        \Omega(\text{W-MSA}) &= 4HWD^2 + 2M^2HWD, \\ 
        \Omega(\text{FFN}) &= 4LD^2, 
\label{complexity_analysis}
\end{align}
where \(L=H\times W\) denotes the sequence length, \(N\) is the SSM state dimension that encodes the context of each channel independently, and \(M\) is the fixed window size. Since SS2D computes four SSM processes corresponding to four scan paths, its complexity can be approximated with \(4\Omega(\text{SSM})\). For SSM, \(3L(2D)N\) comes from the computation of \(\Bar{B}, C\), and \(\Bar{D}\) while the computation of \(\Bar{A}\) contributes to the complexity of \(L(2D)N\). As a result, the overall complexity of VSSM can be approximated as \(\Omega(\text{SS2D})\), which is linear with respect to \(L\). Given a window with fixed size \(M=8\), the complexity of W-MSA is linear with respect to sequence length.
Hence, all three components of LaMamba, namely VSSM, local attention, and FFN, have linear complexity. Built upon LaMamba, LaMamba-Diff is naturally linear. We note that the following operations contribute negligible GFLOPs and are excluded from this analysis for simplicity: reshaping and merging the 4 SSM outputs in SS2D; AdaLN and scaling multiplications with \(\alpha\) in LaMamba blocks.

\begin{table}[t]
\centering
\scalebox{0.8}{
\begin{tabular}{cccc} 
\toprule
Resolution &Model  & Sequence length & \# GFLOPs \\
\hline
\hline
\multirow{2}{*}{\(256\times 256\)} &DiT-XL/2 & 256 & 118.64 \\
&LaMamba-Diff-XL & 1024 & 50.46 \\
\midrule
\multirow{2}{*}{\(512\times 512\)} &DiT-XL/2 & 1024 & 524.60 \\
&LaMamba-Diff-XL & 4096 & 201.20  \\
\midrule
\multirow{2}{*}{\(1024\times 1024\)} &DiT-XL/2 & 4096 & 2910.30\\ 
&LaMamba-Diff-XL & 16384 & 804.18  \\

\bottomrule
\end{tabular}}
\caption{Comparison of GFLOPs between LaMamba-Diff-XL and DiT-XL/2 for different image resolutions.}
\label{computation_analysis}
\end{table}

Empirically, we compare GFLOPs of LaMamba-Diff-XL with DiT-XL/2 for different image resolutions in Table~\ref{method_computation_analysis}. With a comparable number of parameters, LaMamba-Diff-XL incurs 57.5\% and 72.4\% less GFLOPs at \(256\times 256\) and \(1024\times 1024\) resolutions, respectively.
Notably, LaMamba-Diff's U-Net architecture decreases computational complexity by downsampling the input sequence with patch merging in two encoder stages.

\section{Experiments}

\subsection{Setting}

\noindent\textbf{Latent diffusion model.} For all experiments, LaMamba-Diff operates in the latent space of an off-the-shelf pre-trained VAE~\footnote{https://huggingface.co/stabilityai/sd-vae-ft-mse} from Stable Diffusion~\cite{rombach2022high}, which has a downsampling factor of 8. For instance, an RGB image \(x_0\) of size \({256\times 256\times 3}\) would be encoded into a compressed latent \(z_0 = \mathcal{E}(x_0)\) of size \({32\times 32\times 4}\). Our LaMamba-Diff is trained to learn the reverse diffusion process in this \(\mathcal{Z}\)-space using diffusion hyperparameters from ADM~\cite{dhariwal2021diffusion}. Specifically, we follow their embedding approach to encode timestep and class label conditions, and use a linear variance scheduler and ADM's parameterization of covariance \(\Sigma_{\theta}\). Images are generated by sampling novel latents \(z\) and passing them to the VAE decoder \(x=\mathcal{D}(z)\).

\noindent\textbf{Training Details.} We follow the training and hyperprameter settings in \cite{peebles2023scalable} to train variants of LaMamba-Diff on the ImageNet dataset~\cite{russakovsky2015imagenet} at \(256 \times 256\) and \(512 \times 512\) resolutions using classifier free guidance~\cite{ho2022classifier}. Specifically, we utilize the weight initialization techniques in \cite{peebles2023scalable} and employ the AdamW optimizer~\cite{loshchilov2017decoupled} with a constant learning rate of 1e-4, a global batch size of 256, and no weight decay. An exponential moving average (EMA) of LaMamba-Diff weights is maintained during training with a decay of 0.9999. We follow the setting in \cite{liu2024vmamba} for VSSM hyperparameters.

\begin{table*}
    \centering
    \scalebox{0.9}{
    \begin{tabular}{l|ccc|ccccc}
    \toprule
    \textbf{ImageNet 256×256} \\
    \midrule
    \midrule
    Model & Parameters(M) & GFLOPs & Training Steps(M) & FID$\downarrow$ & sFID$\downarrow$ & IS$\uparrow$ & Precision$\uparrow$ & Recall$\uparrow$\\
    \midrule
         \textcolor{lightgray}{BigGAN-deep~\cite{brock2018large}} & \textcolor{lightgray}{} & \textcolor{lightgray}{} & \textcolor{lightgray}{}& \textcolor{lightgray}{6.95}& \textcolor{lightgray}{7.36}& \textcolor{lightgray}{171.40}& \textcolor{lightgray}{0.87}& \textcolor{lightgray}{0.28}\\
     \textcolor{lightgray}{StyleGAN-XL~\cite{sauer2022stylegan}} & \textcolor{lightgray}{} &\textcolor{lightgray}{}& \textcolor{lightgray}{} & \textcolor{lightgray}{2.30}& \textcolor{lightgray}{4.02}& \textcolor{lightgray}{265.12}& \textcolor{lightgray}{0.78}& \textcolor{lightgray}{0.53}\\
     \midrule
    \rowcolor{ggray} &&& \multicolumn{5}{c}{Unconditional} \\
    
     ADM~\cite{dhariwal2021diffusion} & 554 &1120 & 2.0 &10.94& 6.02 &100.98& 0.69& 0.63\\
     ADM-U~\cite{dhariwal2021diffusion} & 608 &742& 2.0 &7.49 &5.13& 127.49& 0.72& 0.63\\
     LDM-8~\cite{rombach2022high} &395 &79 & 1.2 &15.51& -& 79.03& 0.65& 0.63\\
     LDM-4~\cite{rombach2022high} & 391 &103 & 0.8 &10.56& -& 103.49& \textbf{0.71}& 0.62\\
     DiT-XL/2~\cite{peebles2023scalable}&675& 118 & 7.0 & 9.62& 6.85& 121.50& 0.67& \textbf{0.67} \\
     DiffuSSM-XL~\cite{yan2023diffussm} &673& 280 & 2.6 & 9.07& \textbf{5.52}& 118.32& 0.69& 0.64 \\
     \textbf{LaMamba-Diff-XL} & 656 & 50 & 2.0 &\textbf{6.12} & 5.58 & \textbf{149.13} & \textbf{0.71} & 0.64 \\
    \rowcolor{ggray} &&& \multicolumn{5}{c}{Classifier-free guidance} \\
     ADM-G, ADM-U~\cite{dhariwal2021diffusion} & 673 & 761 & 2.0 & 3.94& 6.14 &215.84& 0.83& 0.53\\
     LDM-8-G~\cite{rombach2022high} &395&79 & 1.2 & 7.76& -& 209.52& 0.84& 0.35\\
     LDM-4-G~\cite{rombach2022high} &391 &103 & 0.8 & 3.60& -& 247.67& \textbf{0.87}& 0.48\\
     U-ViT-H/2-G~\cite{bao2023all} &501 & 133 & 2.0 & 2.29& -& 247.67& \textbf{0.87}& 0.48 \\
     DiffuSSM-XL-G~\cite{yan2023diffussm} &673& 280 & 2.6& 2.28& 4.49& 259.13& 0.86& 0.56 \\
     DiT-XL/2-G~\cite{peebles2023scalable}&675& 118 & 7.0 & 2.27&  4.60&  278.24&  0.83&  \textbf{0.57} \\
     \textbf{LaMamba-Diff-XL-G} &656 & 50 & 2.0 & \textbf{2.04}& \textbf{4.52} & \textbf{296.07} & 0.84 & 0.56 \\

    \midrule
    \textbf{ImageNet 512×512} \\
    \midrule
    \midrule
    \rowcolor{ggray} &&& \multicolumn{5}{c}{Unconditional} \\
    DiT-XL/2~\cite{peebles2023scalable}&675& 524 &3.0 & 12.03 & 7.12 & 105.25 & 0.75 & \textbf{0.64} \\
    \textbf{LaMamba-Diff-XL} & 656 & 201 & 3.0 & \textbf{7.76} & \textbf{6.89} & \textbf{122.41} & \textbf{0.81} & 0.60 \\
    \rowcolor{ggray} &&& \multicolumn{5}{c}{Classifier-free guidance} \\
     ADM-G, ADM-U~\cite{dhariwal2021diffusion} & 774 & 2834 &1.9 &3.85& 5.86 &221.72& 0.84& 0.53\\
     U-ViT-H/4-G~\cite{bao2023all} &501 &133 & 2.0& 4.05& 8.44& 261.13& 0.84& 0.48 \\
     DiffuSSM-XL-G~\cite{yan2023diffussm} &673& 1066 & 1.2 & 3.41& 5.84 &255.06& 0.85& 0.49 \\
     DiT-XL/2-G~\cite{peebles2023scalable}&675& 524 &3.0 & 3.04& \textbf{5.02}& 240.82& 0.84 &\textbf{0.54} \\
     \textbf{LaMamba-Diff-XL-G} & 656 & 201 & 3.0 & \textbf{3.01} & 5.15 & \textbf{277.19} & 0.85 & 0.52 \\
    \bottomrule
    \end{tabular}}
    \caption{\textbf{Quantitative comparison of image generation quality on ImageNet 256x256 and 512x512.} We denote classifier-free guidance by appending "-G" to model names. LaMamba-Diff-XL achieves competitive performance with substantially fewer GFLOPs. The training steps denote the number of iterations trained using a batch size of 256.} 
    \label{main_imagenet}
\end{table*}

\noindent\textbf{Evaluation.} 
We measure model performance with Fréchet Inception Distance (FID)~\cite{heusel2017gans}, a standard metric for assessing the quality of generated images. For a fair comparison with prior works, we use 250 DDPM sampling steps to sample 50K images to compute FID~\cite{peebles2023scalable} using the ADM's TensorFlow implementation~\cite{dhariwal2021diffusion}. We also report sFID~\cite{nash2021generating}, Inception Score~\cite{salimans2016improved}, and Precision and Recall~\cite{kynkaanniemi2019improved} as secondary performance metrics.

\subsection{Main Results}

\begin{figure}[h]
\centering
         \includegraphics[width=\columnwidth]{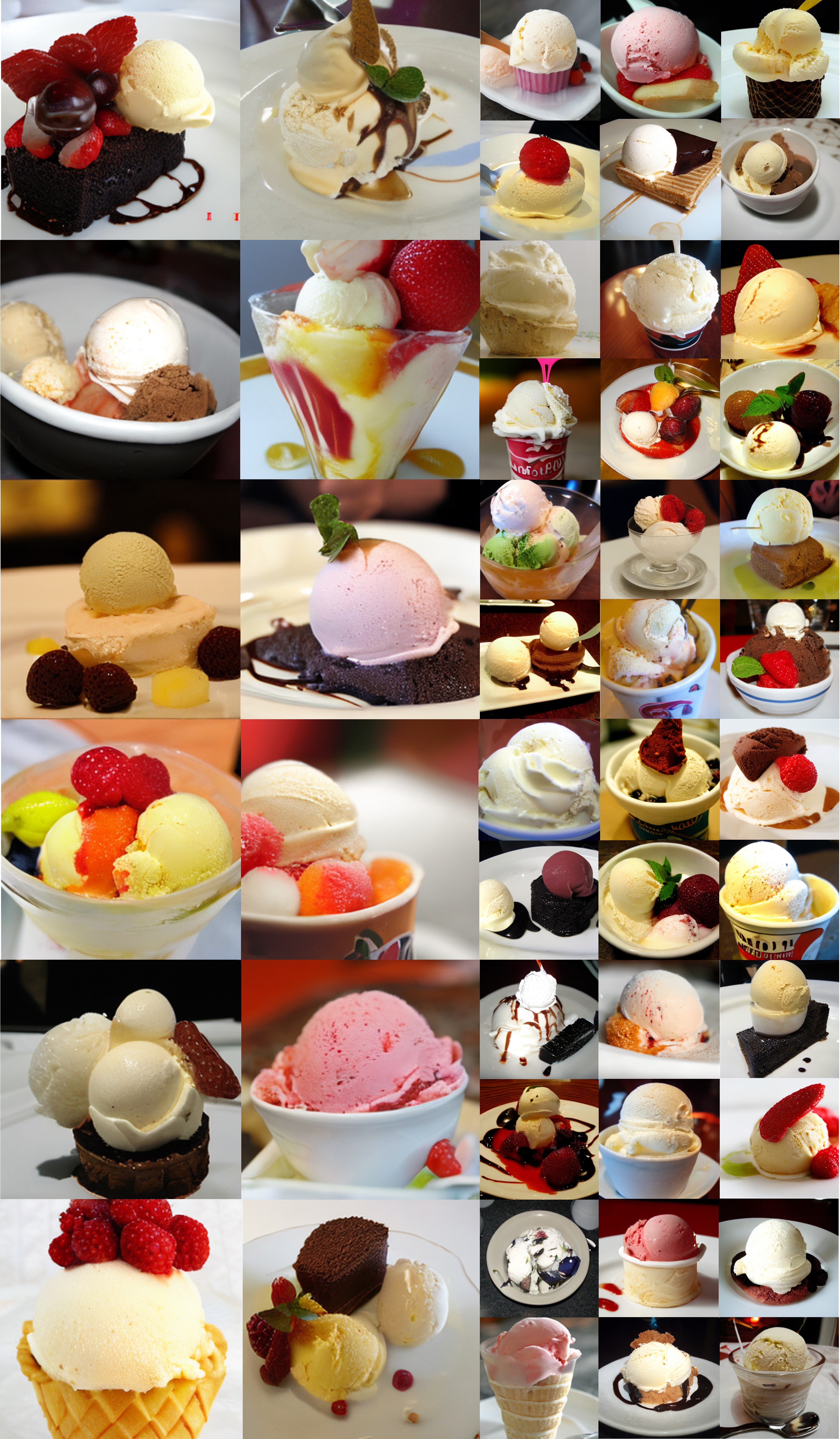}
         \caption{ImageNet \(256 \times 256\) samples generated by LaMamba-Diff-XL using a classifier-free guidance scale of 4.0. \\ Class: Ice Cream}
         \label{sample_ice_cream}
\end{figure}

\begin{figure}[t]

 \centering
 \includegraphics[width=\columnwidth]{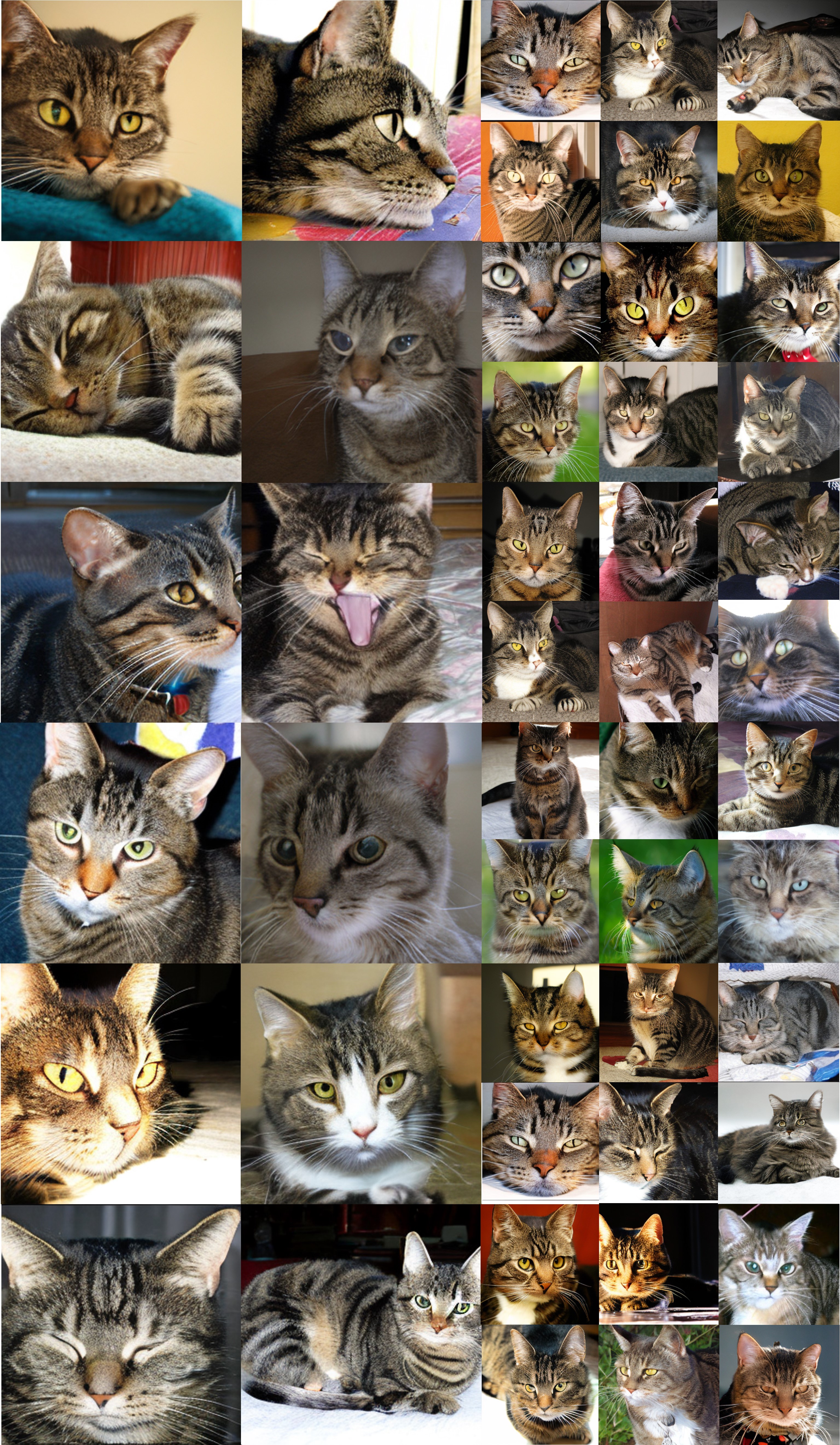}
 \caption{ImageNet \(256 \times 256\) samples generated by LaMamba-Diff-XL using a classifier-free guidance scale of 4.0. \\ Class: Tabby Cat}
 \label{sample_tabby_cat}
\end{figure}

\begin{figure}[t]
\centering
         \includegraphics[width=\columnwidth]{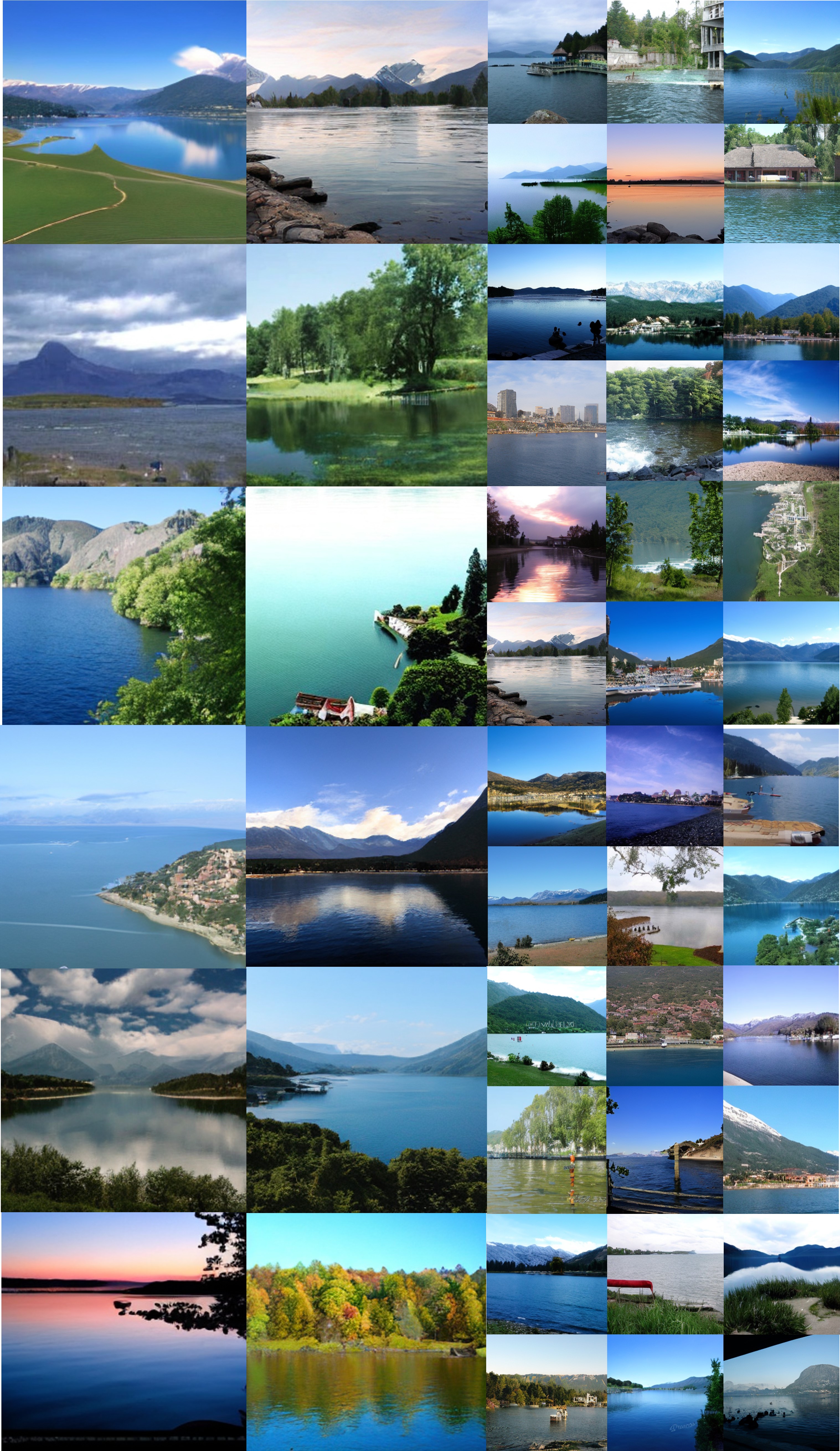}
         \caption{ImageNet \(256 \times 256\) samples generated by LaMamba-Diff-XL using a classifier-free guidance scale of 2.0. \\ Class: Lakeshore}
         \label{sample_lakeshore}
\end{figure}

\begin{figure}[t]

 \centering
 \includegraphics[width=\columnwidth]{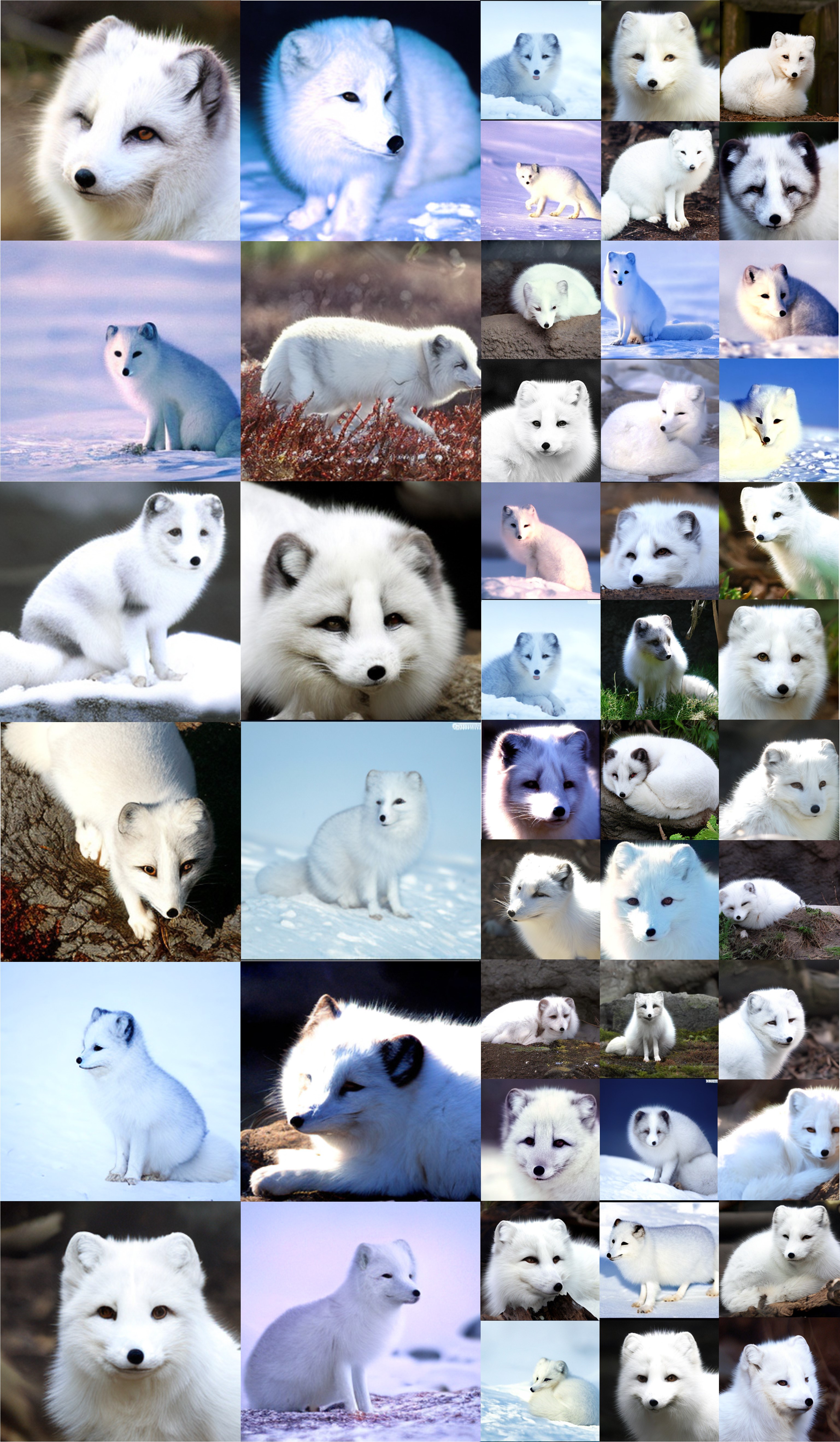}
 \caption{ImageNet \(256 \times 256\) samples generated by LaMamba-Diff-XL using a classifier-free guidance scale of 2.0. \\ Class: Arctic Fox}
 \label{sample_arctic_fox}
\end{figure}

\begin{figure}[t]
\centering
         \includegraphics[width=\columnwidth]{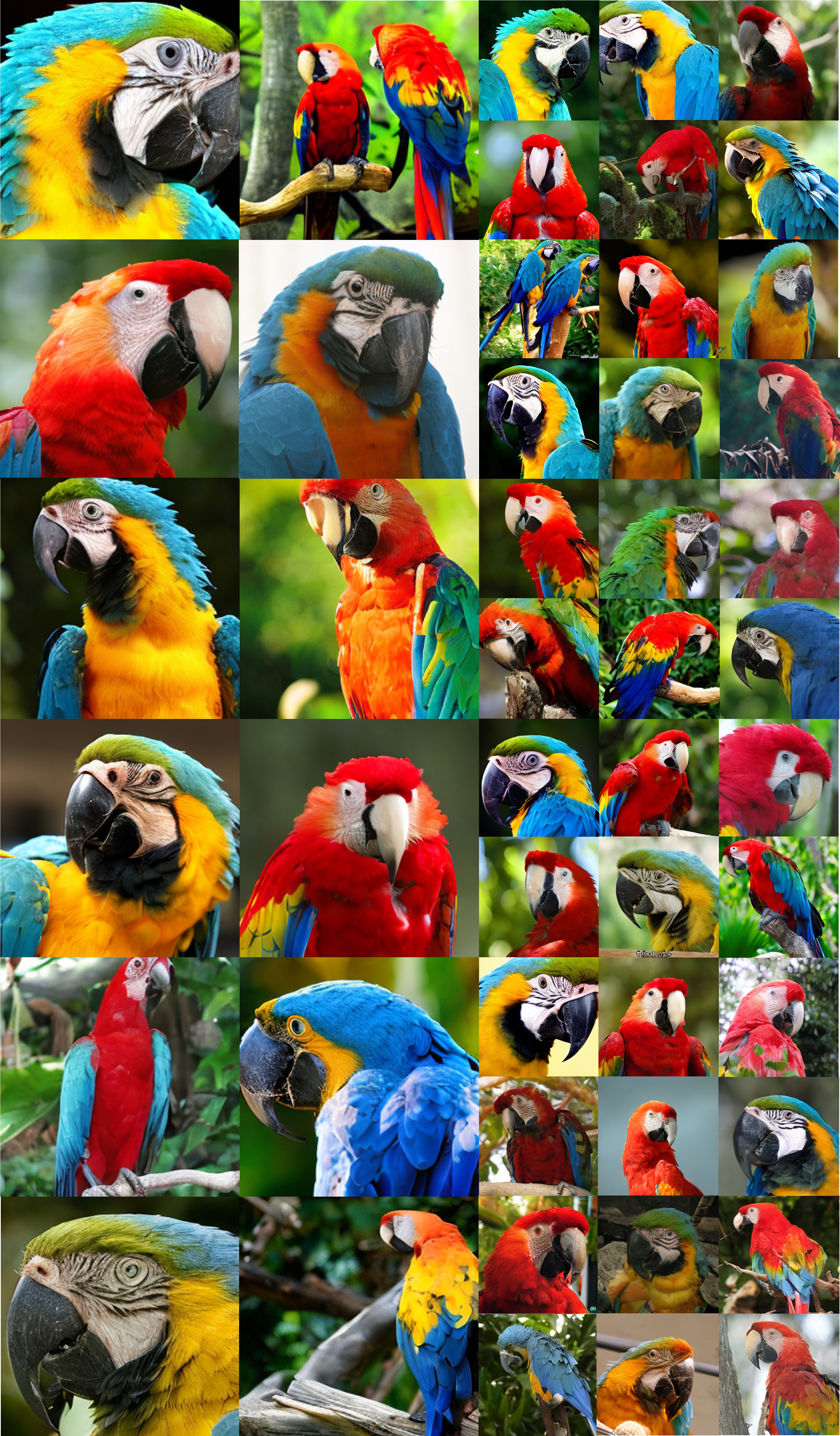}
         \caption{ImageNet \(256 \times 256\) samples generated by LaMamba-Diff-XL using a classifier-free guidance scale of 1.5. \\ Class: Macaw}
         \label{sample_macaw}
\end{figure}

\begin{figure}[t]

 \centering
 \includegraphics[width=\columnwidth]{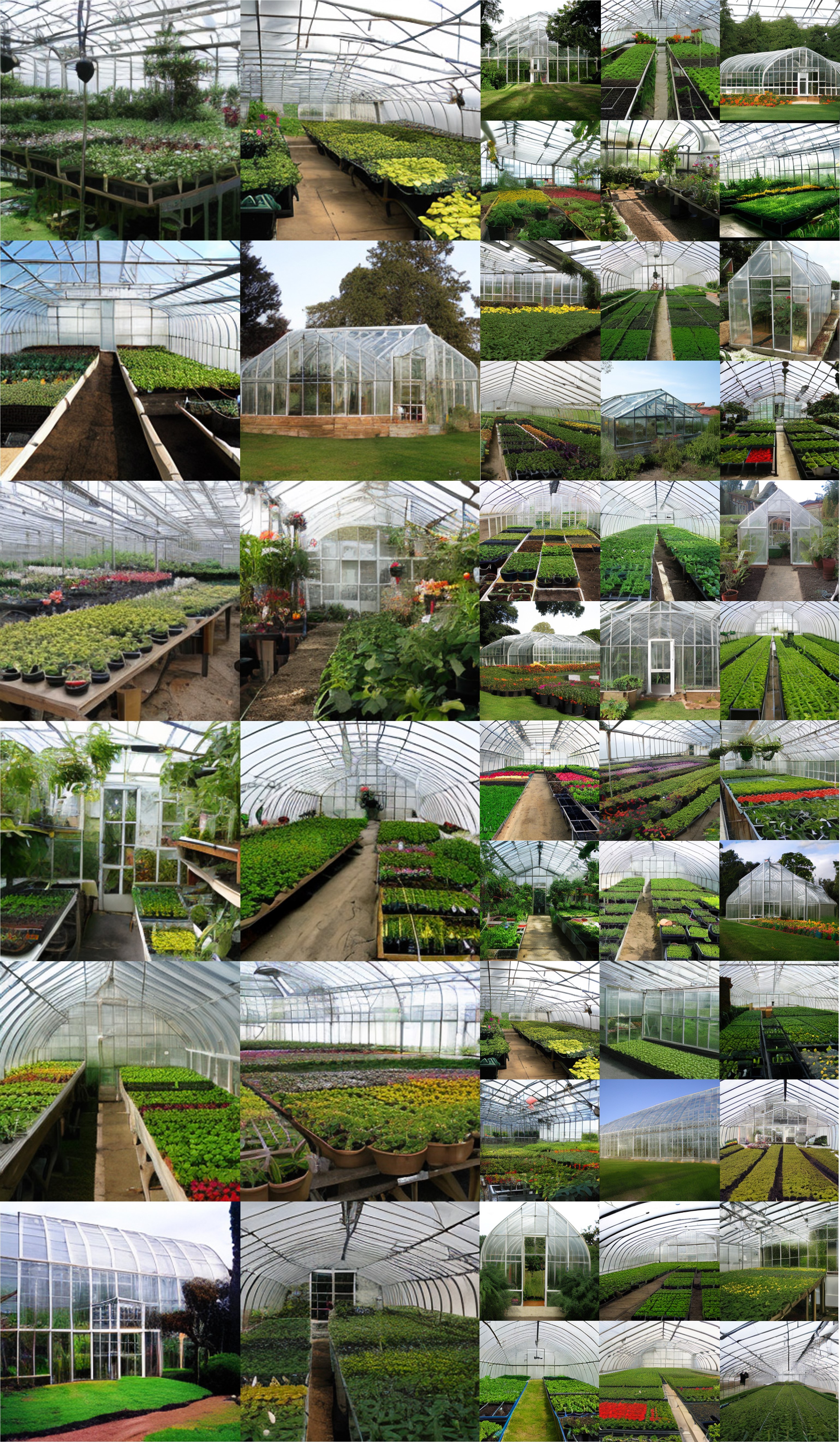}
 \caption{ImageNet \(256 \times 256\) samples generated by LaMamba-Diff-XL using a classifier-free guidance scale of 1.5. \\ Class: Greenhouse}
 \label{sample_greenhouse}
\end{figure}

\textbf{ImageNet 256x256}. We compare our largest model, LaMamba-Diff-XL trained for 2M steps, with state-of-the-art (SOTA) diffusion models on ImageNet at 256x256 resolution. Table~\ref{main_imagenet} reports quantitative performance evaluation results, while qualitative examples are provided in Figures~\ref{sample_ice_cream}-\ref{sample_greenhouse}. 
In both unconditional and class-conditional settings, LaMamba-Diff-XL outperforms attention-only (DiT~\cite{peebles2023scalable},U-ViT~\cite{bao2023all}) and SSM-only (DiffuSSM~\cite{yan2023diffussm}) methods, achieving clearly lower FID scores. 
More importantly, LaMamba-Diff-XL achieves SOTA FIDs while requiring substantially fewer GFLOPs. 
In unconditional image generation, LaMamba-Diff-XL achieves a SOTA FID of 6.12, improving the previous SOTA FID (9.07) by 2.95 (33\%) with 82\% fewer GFLOPs and 17M fewer parameters. Furthermore, LaMamba-Diff-XL achieves the highest Inception Score (IS) and Precision.
In conditional image generation with classifier-free guidance (CFG)~\cite{ho2022classifier}, LaMamba-Diff-XL achieves a SOTA FID of 2.04 using 57.6\% fewer GFLOPs and 19M fewer parameters compared to DiT-XL/2.
Moreover, LaMamba-Diff-XL achieves the highest Inception Score of 296.07.
Note that the previous SOTA conditional FID of 2.10 was achieved by DiS-H/2~\cite{fei2024DiS} (900M), and LaMamba-Diff-XL uses 27\% fewer parameters and needs 77\% fewer GFLOPS.

\begin{figure}[t]
\centering
         \includegraphics[width=\columnwidth]{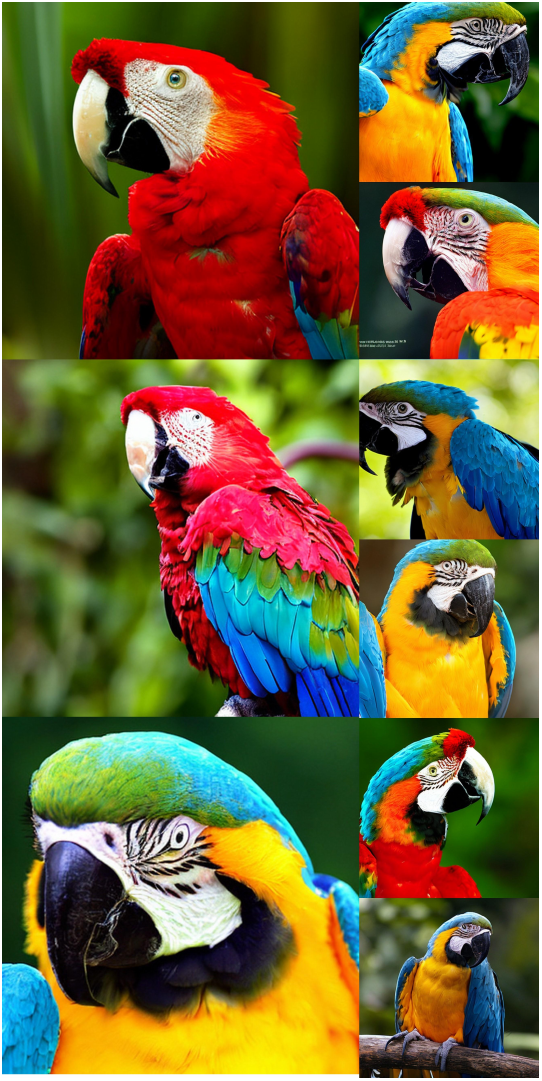}
         \caption{ImageNet \(512 \times 512\) samples generated by LaMamba-Diff-XL using a classifier-free guidance scale of 4.0. \\ Class: Macaw}
         \label{sample_macaw_512}
\end{figure}

\begin{figure}[t]

 \centering
 \includegraphics[width=\columnwidth]{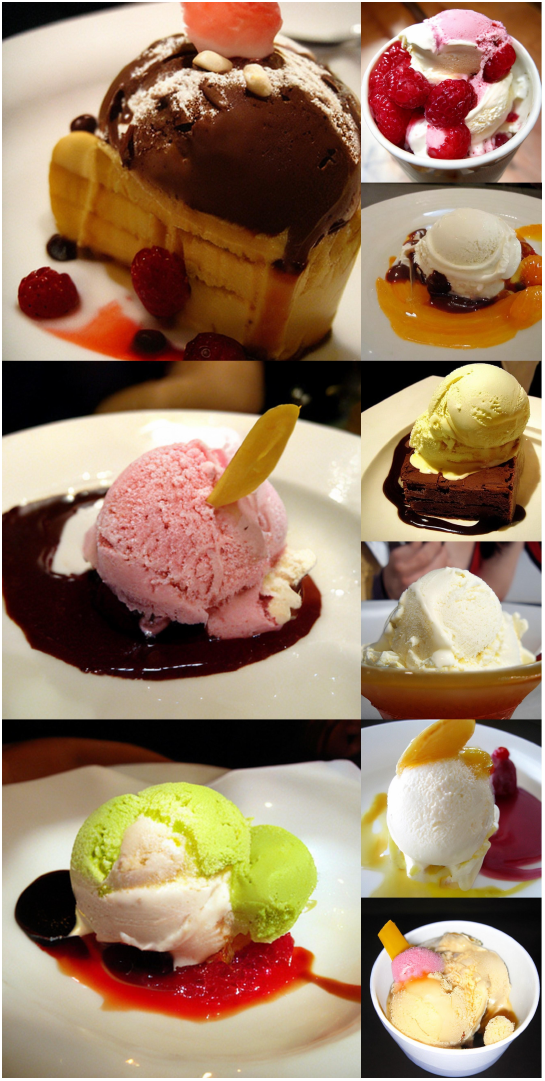}
 \caption{ImageNet \(512 \times 512\) samples generated by LaMamba-Diff-XL using a classifier-free guidance scale of 4.0. \\ Class: Ice Cream}
 \label{sample_ice_cream_512}
\end{figure}

\begin{figure}[t]
\centering
         \includegraphics[width=\columnwidth]{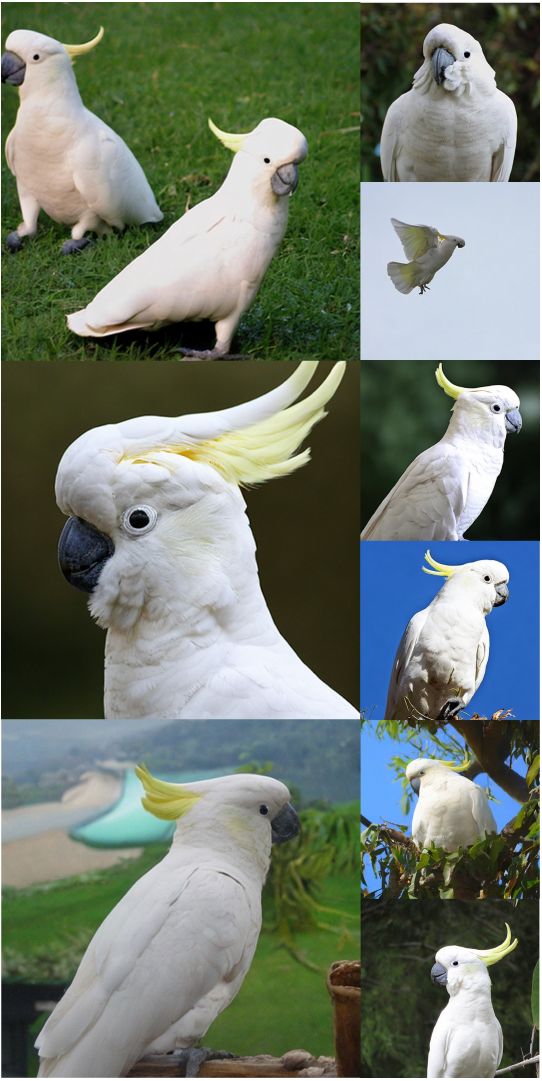}
         \caption{ImageNet \(512 \times 512\) samples generated by LaMamba-Diff-XL using a classifier-free guidance scale of 2.0. \\ Class: Sulphur-crested Cockatoo}
         \label{sample_Sulphur-crested_Cockatoo_512}
\end{figure}

\begin{figure}[t]

 \centering
 \includegraphics[width=\columnwidth]{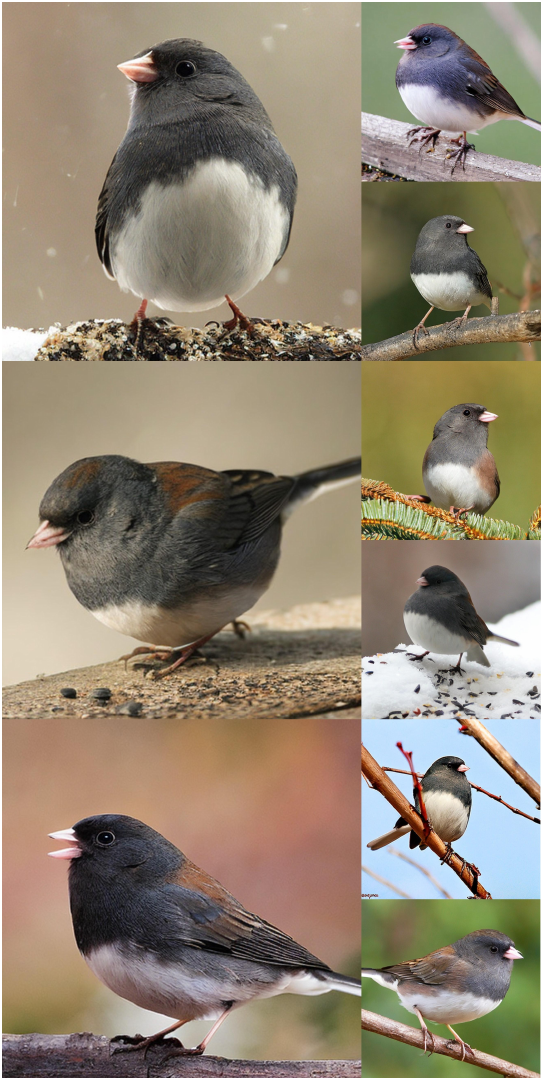}
 \caption{ImageNet \(512 \times 512\) samples generated by LaMamba-Diff-XL using a classifier-free guidance scale of 2.0. \\ Class: Tenco snowbird}
 \label{sample_Tenco_snowbird_512}
\end{figure}

\begin{figure}[t]
\centering
         \includegraphics[width=\columnwidth]{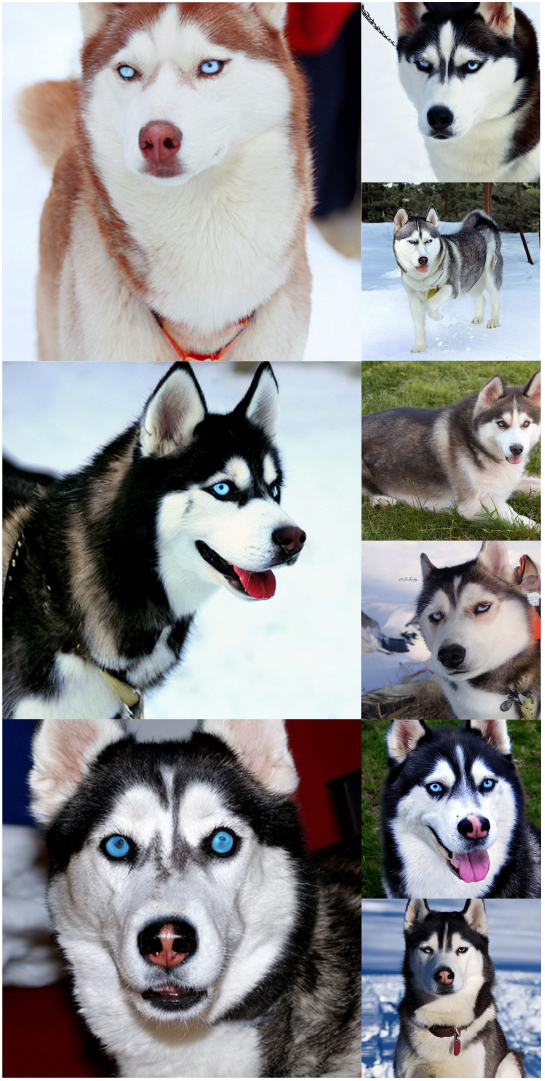}
         \caption{ImageNet \(512 \times 512\) samples generated by LaMamba-Diff-XL using a classifier-free guidance scale of 1.5. \\ Class: Husky}
         \label{sample_Sulphur-sample_Husky_512}
\end{figure}

\begin{figure}[t]

 \centering
 \includegraphics[width=\columnwidth]{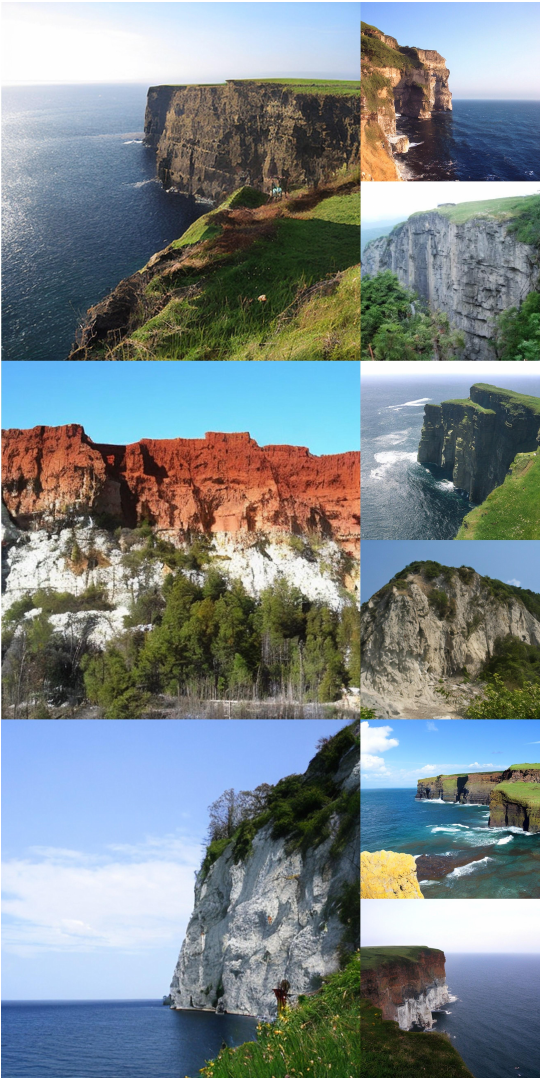}
 \caption{ImageNet \(512 \times 512\) samples generated by LaMamba-Diff-XL using a classifier-free guidance scale of 1.5. \\ Class: Cliff}
 \label{sample_cliff_512}
\end{figure}

\textbf{ImageNet 512x512.} We evaluate the image generation quality of LaMamba-Diff-XL on ImageNet at 512x512 resolution.
Here, LaMamba-Diff-XL processes 4096 tokens with an input latent of size \(64\times 64 \times 4\). 
Table~\ref{main_imagenet} shows comparisons with SOTA methods, while Figures~\ref{sample_macaw_512}-\ref{sample_cliff_512} exhibits examples of generated images.
LaMamba-Diff-XL achieves lower FID compared to all prior diffusion models with similar sizes and uses 61.6\% less GFLOPs compared to DiT-XL/2.
In the unconditional setting, LaMamba-Diff-XL substantially improves the FID of DiT-XL/2 from 12.03 to 7.76.
When using classifier-free guidance, LaMamba-Diff-XL achieves a SOTA FID of 3.01 and the highest IS of 277.19.
Note that DiS-H/2~\cite{fei2024DiS} achieves a slightly lower FID (2.88) using 37\% more parameters and 258\% more GFLOPs.
These results highlight the capability and efficiency of LaMamba-Diff in high-resolution image generation.

\begin{table}[t]
    \centering
    \scalebox{0.9}{
    \begin{tabular}{lccc}
    \toprule
        Model & Parameters (M) & GFLOPs  & FID \\
    \midrule
    \midrule
        LaMamba-Diff-S & 32 & 3.19 & 55.68 \\
        LaMamba-Diff-B & 127 & 12.32 & 29.71 \\
        LaMamba-Diff-L & 449 & 33.39 &  18.15 \\
        LaMamba-Diff-XL & 656 & 49.90 & 12.86 \\

        \midrule
        DiT-S/2 & 33 & 6.06 & 68.40 \\
        DiT-B/2 & 130 & 23.01 & 43.47 \\
        DiT-L/2 & 458 & 80.71 & 23.33 \\
        DiT-XL/2 & 675 & 118.64 & 19.47 \\
    
    \bottomrule
    \end{tabular}
    }
    \caption{Comparison of LaMamba-Diff and DiT scaling properties based on unconditional FID and models trained on ImageNet 256x256 for 400k steps using a batch size 256.}
    \label{experiment_scaling}
\end{table}

\textbf{Scaling model complexity}
Here we evaluate the scaling properties of LaMamba-Diff by training the four models presented in Table~\ref{method_architecture} on ImageNet 256x256 for 400k steps using the same hyperparameters. Figure~\ref{intro_FID_comparisons} (left) and Table~\ref{experiment_scaling} present our results. We observe substantial FID improvements when increasing the number of parameters. In comparison to LaMamba-Diff-S with 32M parameters, LaMamba-Diff-XL with 656M parameters improves the FID score by 42.82.
LaMamba-Diff models outperform DiT models with a comparable number of parameters by consistently exhibiting superior performance while using significantly fewer GFLOPs. These observations underscore the superior scaling capability of LaMamba-Diff models.

\subsection{Ablation studies}
In this section, we investigate the effectiveness of each proposed component in our LaMamba block as well as in the overall network architecture. Unless otherwise stated, we report FID-50K for variants of LaMamba-Diff Tiny trained on ImageNet 256x256 for 400K steps with a batch size of 256.

\begin{table}
    \centering
    \scalebox{0.9}{
    \begin{tabular}{lcc}
    \toprule
        Method & GFLOPs & FID \\
    \midrule
    \midrule
        LaMamba-Diff & 3.19 & 55.68 \\
        w/o attention & 2.45 & 69.78 \\
        w/o shifting & 3.19 & 58.81 \\
    \midrule
        global attention & 4.24 & 53.74 \\

    \bottomrule
    \end{tabular}}
    \caption{Ablation study of attention in LaMamba blocks.\vspace{-3mm}}
    \label{ablation_attention}
\end{table}

\textbf{Local attention.} The local attention in our LaMamba block is crucial to success, complementing VSSM by capturing detailed local contextual information. In Table~\ref{ablation_attention}, we study the effect of local attention and different attention strategies. Specifically, we removed the local attention module entirely (w/o attention) or disabled the alternating window shift (w/o shifting). 
Additionally, we replaced local attention with global attention.

Table~\ref{ablation_attention} presents our results, indicating that removing local attention leads to a substantial increase in FID by 14.1, confirming the importance of modeling detailed local contexts.
Additionally, removing the shifting scheme for windowed attention leads to a 3.13 increase in FID, suggesting that the propagation of fine-grained information across window boundaries is useful.
Global attention marginally improves FID by 3.5 despite bringing a significant 32.5\% computational overhead, indicating that the combination of local attention and VSSM is as powerful as global attention plus VSSM in capturing both global and local contexts while incurring much less computation.

\textbf{Visual state space module (VSSM).} In LaMamba blocks, VSSM is responsible for capturing global contexts and 2D spatial information from inputs. Here, we explore the performance of different VSSMs discussed in Section~\ref{method_LaMamba}. We follow the original implementation and Mamba hyperparameters released in the corresponding GitHub repositories. As demonstrated in Table~\ref{ablation_mamba}, LaMamba-Diff with spatially continuous SS2D achieves superior performance, and spatially continuous scanning~\cite{yang2024plainmamba} improves FID by 0.89. 
Under comparable numbers of parameters, SS2D outperforms other VSSM variants significantly, thereby highlighting its effectiveness in capturing global contexts in latent diffusion models.

\begin{table}[t]
    \centering
    \scalebox{0.8}{
    \begin{tabular}{lccc}
    \toprule
        Method & Parameters (M) & GFLOPs  & FID \\
    \midrule
    \midrule
        LaMamba-Diff & 32 & 3.19 & 55.68 \\
        SS2D~\cite{liu2024vmamba} & 32 & 3.19 & 56.57 \\
        Bi-directional Mamba~\cite{zhu2024vim} & 30 & 2.62 & 83.99 \\
        LocalVMamba~\cite{huang2024localmamba} & 32 & 3.32 & 92.10 \\
        EfficientVMamba~\cite{pei2024efficientvmamba} & 31 & 2.40 & 105.93 \\
    
    \bottomrule
    \end{tabular}
    }
    \caption{Ablation study of VSSM in LaMamba blocks.}
    \label{ablation_mamba}
\end{table}

\begin{table*}[t]
    \centering
    \scalebox{0.85}{
    \begin{tabular}{llcccc} 
    \toprule
    Ablation Type & Method & Initial hidden dim & Parameters (M) & GFLOPs & FID\\
    \hline
    \hline
    & LaMamba-Diff & 96 & 32 & 3.19  & 55.68\\
    
    \midrule
    \multirow{2}{*}{Downsampling} & 1x & 144& 30 & 6.99 & 53.49 \\
     & 3x & 48 & 33 & 1.93 & 62.23\\

    \midrule
    \multirow{3}{*}{Block Number} & Decoder -1 & 104 &32& 3.08 & 54.93 \\
    & All +1 & 80 & 33& 3.51 & 63.54 \\
    & All -1 & 112 & 35& 2.93& 58.17 \\
     
    \midrule
    \multirow{4}{*}{Isotropic architecture} & Patch size 1 & 384 & 34 & 24.96 & 52.01\\
    & Patch size 2 & 384 & 34 & 6.19 & 91.89\\
    & Patch size 4 & 384 & 34 & 1.96 & 129.92 \\
    & Patch size 8 & 384 & 34 & 0.65 & 172.10 \\
    \midrule
    U-Net Shortcut& Concatenate & 64 & 33 & 3.02 & 89.10 \\

    \bottomrule
    \end{tabular}}
    \caption{Ablation studies on architectural design choices.}
    \label{ablation_architecture}
\end{table*}

\textbf{Network architecture.}
In this section, we conduct a comparative analysis of various architectural design choices. Specifically, we evaluate four types of architectural variations: the adoption of an isotropic architecture instead of a U-Net architecture, the number of downsampling blocks in U-Net, the number of LaMamba blocks in each U-Net stage, and the adoption of channel concatenation instead of addition in U-Net shortcuts. The hidden dimension is adjusted to ensure that all architectural variants considered in the same comparison have a similar number of parameters. The results of these ablations are presented in Table~\ref{ablation_architecture}.

In LaMamba-Diff, we downsample twice, resulting in a bottleneck resolution of \(8\times 8\) for \(256\times 256\) input images. Here, we investigate the impact of downsampling only once (\(1\times\)) and three times (\(3\times\)), corresponding to bottleneck resolutions of \(16\times 16\) and \(4\times 4\), respectively. Our observations reveal that \(3\times\) downsampling leads to a decline in performance, while \(1\times\) downsampling marginally improves FID by 3.9, albeit with an 118\% increase in GFLOPs.

Next, we analyze the effect of varying the number of LaMamba blocks in each stage of the encoder and decoder in LaMamba-Diff, where each encoder stage has 2 blocks and each decoder stage has 3 blocks. We assess performance variations by using 2 blocks only in every decoder stage (Decoder -1), adding one block to every stage in both encoder and decoder (All +1), or removing one block from every stage in both encoder and decoder (All -1). Our results indicate that increasing the number of blocks in every stage at the cost of decreasing hidden dimensions leads to performance drops. 
Moreover, using less LaMamba blocks in the decoder (Decoder -1) achieves comparable FID and GLFOPs, providing an alternative architectural design for LaMamba-Diff, which nevertheless, adopts the decoder design in StableDiffusion~\cite{rombach2022high} to make it stronger.

We also compare LaMamba-Diff with an isotropic architecture with no downsampling, which is essentially a DiT~\cite{peebles2023scalable} with LaMamba blocks. We observe that an isotropic architecture using a patch size of 1 can marginally improve FID by 3.67 at the cost of 7.8 times more GFLOPs, while larger patch sizes result in inferior performance. In contrast, our U-Net architecture presents a significantly superior trade-off between FID and GFLOPs.

Compared to channel concatenation in U-Net shortcuts~\cite{rombach2022high}, LaMamba-Diff achieves substantially lower FID with a comparable number of parameters and a larger hidden dimension of 96. Overall, architecture ablation studies suggest that the size of hidden dimensions is crucial to the success of LaMamba-Diff.

\section{Conclusion}
In this paper, we have proposed LaMamba-Diff, a novel linear-time backbone network for diffusion models. It efficiently captures both global contexts and local dependencies from input tokens. Our experiments demonstrate that LaMamba-Diff, with comparable number of parameters and significantly fewer GLFOPs, achieves very competitive performance against state-of-the-art diffusion backbones. Furthermore, our LaMamba block demonstrates excellent scalability, and like DiT, can be taken as a generic building block for various types of diffusion models, including both text-to-image and text-to-video diffusion models.

{
    \small
    \bibliographystyle{ieeenat_fullname}
    \bibliography{main}
}
\clearpage


\clearpage
\setcounter{page}{1}
\maketitlesupplementary
\section{Additional Qualitative Results}

\vspace{1.05cm}

\end{document}